\documentclass[letterpaper]{article} 
\usepackage{aaai2026}  
\usepackage{times}  
\usepackage{helvet}  
\usepackage{courier}  
\usepackage[hyphens]{url}  
\usepackage{graphicx} 
\usepackage{natbib}  
\usepackage{caption} 
\usepackage{algorithm}
\usepackage{algpseudocode}
\usepackage{newfloat}
\usepackage{listings}
\DeclareCaptionStyle{ruled}{labelfont=normalfont,labelsep=colon,strut=off} 
\floatstyle{ruled}
\newfloat{listing}{tb}{lst}{}
\floatname{listing}{Listing}
\nocopyright

\usepackage{fontawesome5}
\usepackage{amsmath}
\usepackage{makecell}
\usepackage{colortbl} 
\usepackage{arydshln} 
\usepackage{booktabs} 
\usepackage{multirow}
\usepackage{makecell}
\usepackage{graphicx}
\usepackage{listings}
\usepackage{hyperref}
\usepackage{CJKutf8}
\usepackage{booktabs}
\usepackage{multirow}
\usepackage{rotating}
\usepackage{amssymb}
\usepackage{appendix}
\usepackage{caption}
\usepackage[most]{tcolorbox}
\usepackage{longtable}
\usepackage{newfloat}
\usepackage{listings}

\title{TASE: Token Awareness and Structured Evaluation for Multilingual Language Models}
\author{
    Chenzhuo Zhao$^1$\equalcontrib,
    Xinda Wang$^1$\equalcontrib,
    Yue Huang$^1$,
    Junting Lu$^1$,
    Ziqian Liu$^2$
}
\affiliations{
    \textsuperscript{\rm 1}Peking University
    
    \textsuperscript{\rm 2}Independent Researcher
    
    \{cyzcz, nev\_settle,huangyue\_vl,aidan.lew.37\}@stu.pku.edu.cn
    
    liuziqian25@gmail.com}

\usepackage{bibentry}

\begin{document}

\maketitle

\begin{abstract}
While large language models (LLMs) have demonstrated remarkable performance on high-level semantic tasks, they often struggle with fine-grained, token-level understanding and structural reasoning—capabilities that are essential for applications requiring precision and control. We introduce TASE, a comprehensive benchmark designed to evaluate LLMs' ability to perceive and reason about token-level information across languages. TASE covers 10 tasks under two core categories: token awareness and structural understanding, spanning Chinese, English, and Korean, with a 35,927-instance evaluation set and a scalable synthetic data generation pipeline for training. Tasks include character counting, token alignment, syntactic structure parsing, and length constraint satisfaction. We evaluate over 30 leading commercial and open-source LLMs, including O3, Claude 4, Gemini 2.5 Pro, and DeepSeek-R1, and train a custom Qwen2.5-14B model using the GRPO training method. Results show that human performance significantly outpaces current LLMs, revealing persistent weaknesses in token-level reasoning. TASE sheds light on these limitations and provides a new diagnostic lens for future improvements in low-level language understanding and cross-lingual generalization.Our code and dataset are publicly available at \href{URL}{https://github.com/cyzcz/Tase}.
\end{abstract}

\section{Introduction}

Large Language Models (LLMs) have demonstrated impressive capabilities across a wide range of natural language tasks. They excel in high-level semantic understanding such as instruction following, logical reasoning, long-context comprehension, and code generation.\cite{hua2025researchcodebenchbenchmarkingllmsimplementing,kostikova2025lllmsdatadrivensurveyevolving,wan2024efficientlargelanguagemodels} These strengths have driven their adoption in various applications, including conversational agents, educational tools, and problem-solving systems.
\begin{figure}[ht]
  \centering
  \includegraphics[width=\linewidth]{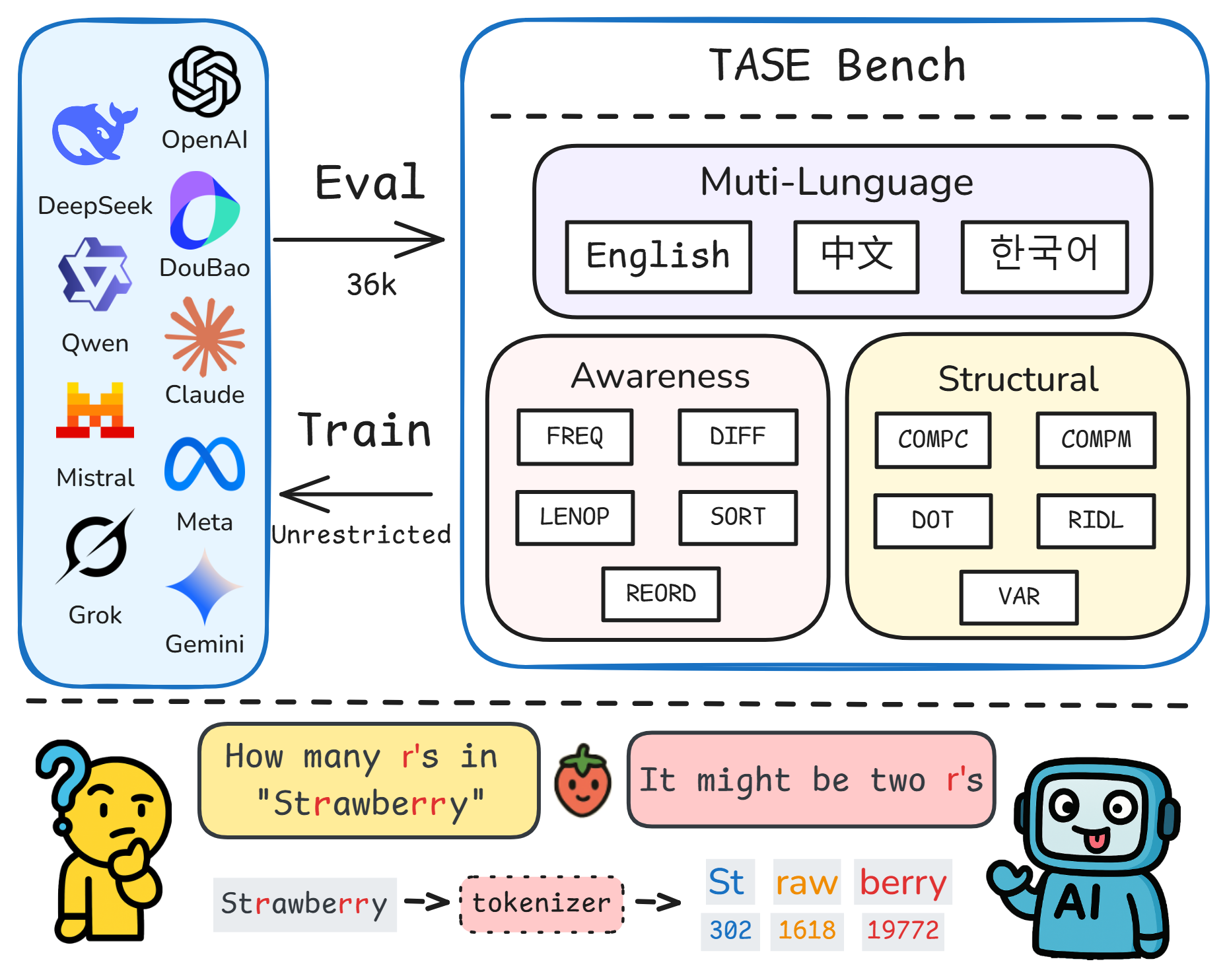}
  \caption{Overview of the TASE benchmark. TASE evaluates LLMs across fine-grained token-level tasks in three languages and two dimensions: token awareness and structural understanding. The strawberry example illustrates common model failures in token-level reasoning.}
  \label{fig:tase-overview}
\end{figure}

Despite their success on complex tasks, LLMs often struggle with surprisingly simple, fine-grained tasks that require token-level perception and structural reasoning.\cite{wang2025tokenizationmattersdegradinglarge,hiraoka2025spellingoutstraightforwardllmscapability} For example, even top-tier models frequently fail to count the number of letter ``r''s in \textit{``strawberry''}, or perform basic operations such as detecting spelling errors or manipulating individual characters. These shortcomings reveal a persistent gap in token awareness—that is, the model's ability to perceive, reason about, and operate on individual tokens or characters with precision.

A key source of this problem is the reliance on subword tokenization schemes like Byte-Pair Encoding \cite{shibata1999byte}, which obscure internal character structures and are not designed for character-level reasoning. This deficiency is especially pronounced in non-English languages like Chinese and Korean, where complex compositional structures pose even greater challenges. Yet, this fundamental blind spot has been largely ignored in mainstream evaluation. Prominent benchmarks like GLUE\cite{wang2018glue}, SuperGLUE\cite{wang2019superglue}, and XNLI\cite{conneau2018xnli} almost exclusively target high-level semantic understanding, overlooking tasks that require direct token manipulation or structural analysis. While recent work has begun to probe these limitations\cite{xu-ma-2024-genius, yehudai-2024-count}, a comprehensive, multilingual benchmark focused on these fine-grained abilities is still critically lacking.

To facilitate a systematic evaluation, we design \textbf{TASE}, a benchmark with ten fine-grained tasks across three typologically diverse languages: English, Chinese, and Korean, representing alphabetic, logographic, and featural writing systems respectively (see Figure~\ref{fig:tase-overview}). TASE evaluates two core dimensions of token-level understanding. The first, token awareness, directly tests a model's perception of linguistic units through tasks such as counting words in a sentence, generating text of a specified length, identifying minimal token differences, and reordering sentences under strict adjacency constraints. The second, structural understanding, probes the model's ability to analyze the internal form of tokens. This includes tasks such as counting characters within a word, solving composition puzzles, reconstructing text from corrupted representations (e.g., dot-matrix patterns), and recognizing visual patterns in text. Together, these tasks evaluate not just symbolic awareness but also structure-sensitive reasoning.

We construct a curated evaluation set of 35{,}928 instances across all tasks and languages, ensuring a broad and reliable basis for performance measurement. Moreover, we develop a scalable synthetic data generation pipeline capable of producing unlimited training examples with guaranteed correctness. This pipeline allows researchers to train or fine-tune models on these tasks and to conduct controlled experiments on how token-level training affects model behavior.

We benchmark more than 30 leading LLMs, including proprietary systems such as GPT-4.1, Claude 4, and Gemini 2.5 Pro, and state-of-the-art open models like DeepSeek-R1. In addition, using our synthetic data and the GRPO algorithm\cite{shao2024deepseekmath}, we fine-tuned a new model based on Qwen2.5-14B-Instruct. Our evaluation reveals a clear and consistent gap between human and model performance. Even the strongest models underperform substantially on TASE tasks, especially those requiring structure-aware reasoning such as visual recognition or component-level decomposition. For example, tasks involving character composition or spatial reasoning often produce incorrect or hallucinated outputs, while simple length-controlled generation is frequently violated. Despite these shortcomings, we find that targeted fine-tuning yields measurable improvements, as our trained model surpasses its base on several tasks, demonstrating that enhancing fine-grained capabilities can help narrow the performance gap. Nevertheless, no model achieves human-level performance across all tasks or languages, underscoring that token-level and structure-aware understanding remains a core challenge for current LLMs.

We summarize our main contributions as follows:
\begin{itemize}
\setlength{\itemsep}{-0.2em}
    \item A multilingual benchmark specifically designed to evaluate token-level awareness and structural reasoning in LLMs, spanning English, Chinese, and Korean.
    \item A reproducible dataset of 35,928 evaluation instances across ten fine-grained tasks, covering both perception and manipulation of linguistic structure.
    \item A scalable synthetic data generation pipeline for each task, enabling training, fine-tuning, and controlled analysis of model behavior.
    \item A comprehensive evaluation of over 20 state-of-the-art LLMs and a fine-tuned 14B custom model, revealing persistent weaknesses and quantifying the gap between models and human performance.
\end{itemize}
\section{Related Work}
\subsection{2.1 High-Level Understanding Benchmarks}

Most existing benchmarks for large language models (LLMs) focus on high-level semantic tasks. GLUE~\cite{wang2018glue} and SuperGLUE~\cite{wang2019superglue} emphasize sentence-level classification, inference, and question answering. Their multilingual extensions, such as XGLUE~\cite{liang2020xglue}, XNLI~\cite{conneau2018xnli}, and TyDiQA~\cite{clark2020tydi}, apply similar task types to non-English languages. More recent efforts like P-MMEval and BenchMAX expand the coverage to multilingual reasoning, coding, and instruction following. However, these benchmarks do not evaluate fine-grained capabilities such as character counting, token alignment, or structural analysis. Furthermore, current benchmarks~\cite{wang2023seaeval,singh2024global,lai2023okapi} lack support for testing token-level or language-agnostic skills, particularly in low-resource settings.

\subsection{2.2 Token Awareness and Fine-Grained Evaluation}

Several studies have recently highlighted that LLMs exhibit surprising failures on basic token-aware tasks. \citet{xu-ma-2024-genius} show that even top-tier models frequently miscount letters within a word, despite understanding their semantic context. \citet{fu2023letter} find that models recognize letters but often fail to count them accurately. \citet{yehudai-2024-count} provide a theoretical framework suggesting that fixed-size transformers struggle with simple counting operations due to architectural limitations. Benchmarks like LMEntry~\cite{efrat2022lmentry} and CUTE test models on elementary character-level operations (e.g., identifying first/last letters, swapping characters) and expose consistent performance gaps. These findings reveal fundamental weaknesses in LLMs’ token-level reasoning, often obscured in high-level evaluations. However, current studies remain fragmented, with evaluations scattered across isolated tasks, and lack a unified or systematic framework for assessing token-level capabilities.

 \begin{figure*}[ht]
  \centering
  \includegraphics[width=\linewidth]{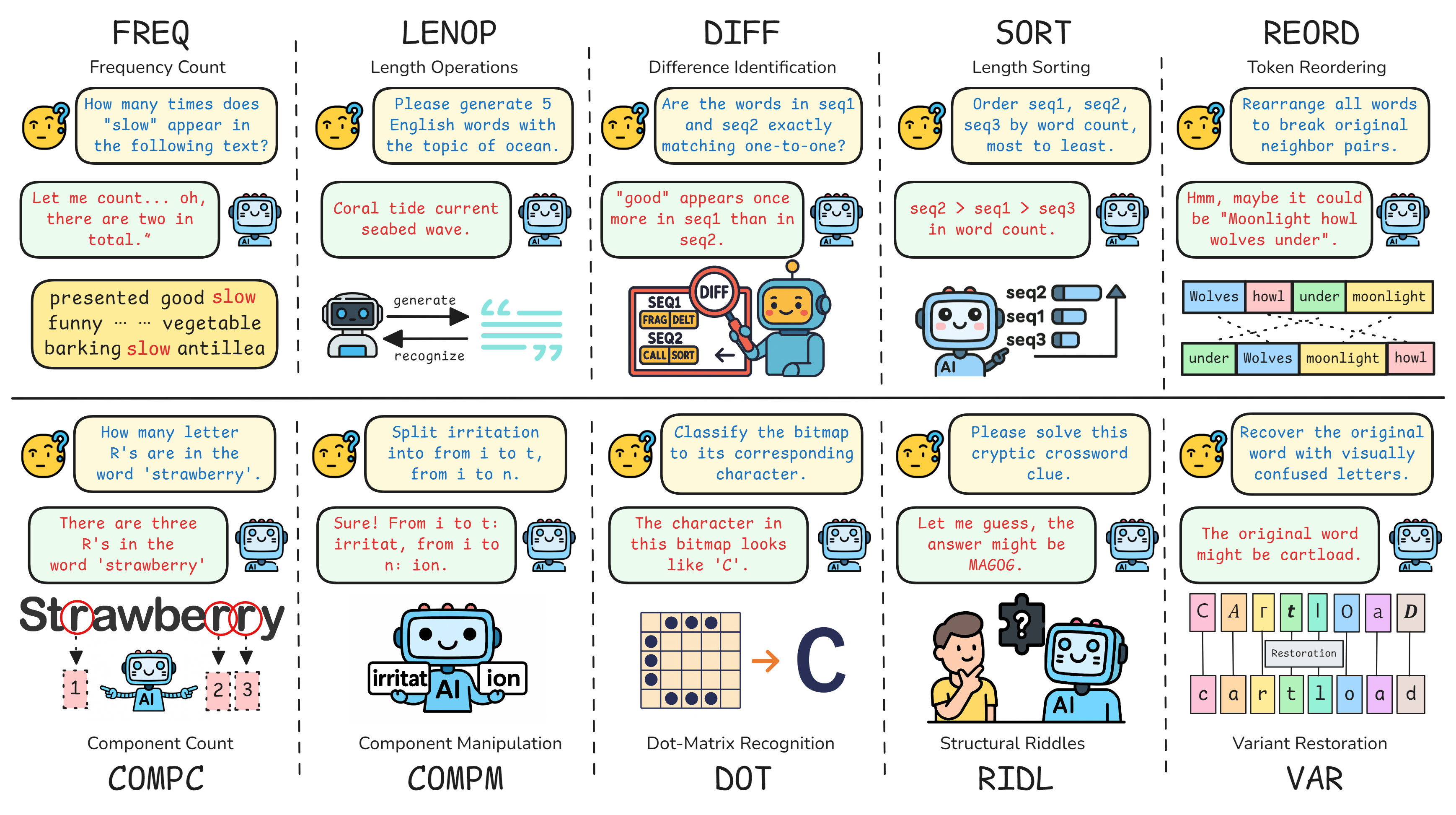}
    \caption{Illustration of the ten TASE tasks, grouped by two core capabilities: \textbf{Token Awareness} (top row) and \textbf{Structural Understanding} (bottom row). Each cell shows an example highlighting the specific reasoning or perception skill evaluated by the task.}
  \label{fig:tase}
\end{figure*}

\subsection{2.3 Structural Reasoning and Text Manipulation}
Beyond token awareness, structural reasoning tasks such as text editing or visual token recognition remain underexplored. CWUM introduces a bilingual benchmark targeting letter-level editing and reordering, reporting large gaps between human and model accuracy. LLMs often fail to enforce formatting constraints, satisfy length conditions, or recover corrupted character structures~\cite{zhou2023instruction}. Attempts to address these issues through synthetic data augmentation or decomposition strategies (for example, spelling out words or breaking tokens into components) offer partial improvements. Still, systematic evaluation suites targeting such structural capabilities across languages remain limited. TASE represents the first systematic multilingual benchmark designed to evaluate both token-level awareness and structural reasoning capabilities.
\section{TASE} 
 \label{sec:benchmark} 
 
 Three characteristics differentiate TASE from existing LLM benchmarks: 
 (1) TASE shifts the evaluation focus from abstract semantic comprehension to concrete, low-level text processing skills, directly probing a model's awareness of tokens and their internal structure. 
 (2) TASE systematically investigates cross-lingual generalization by incorporating parallel tasks across English (alphabetic), Chinese (logographic), and Korean (featural syllabary) to reveal biases associated with different linguistic structures. 
 (3) TASE is designed for scalability and reproducibility, providing not only a fixed evaluation set for fair comparison but also a synthetic data generation pipeline for creating a virtually unlimited volume of training examples. 

 \subsection{3.1 Taxonomy of Language Capabilities} 
 \label{subsec:taxonomy} 
 
We define 10 fundamental low-level language capabilities, organized into two distinct categories: token awareness and structural understanding. These tasks are designed to assess a model's perception and manipulation of the fundamental components of text. A detailed overview and visual examples of each task are provided in Figure~\ref{fig:tase}.
  
 \paragraph{Token Awareness} 
This category focuses on direct operations over discrete token sequences, treating each token as an atomic unit without considering its internal structure.

We formalize all tasks in this category as functions operating over token sequences:
\[
F_{\text{token}} : S \;\text{or}\; (S_1, S_2) \;\text{or}\; \mathcal{S} \rightarrow A
\]
where \( S = \langle t_1, t_2, \dots, t_n \rangle \) is a token sequence, \( \mathcal{S} \) is a set of such sequences, and \( A \) is a scalar, token, or reordered sequence depending on the task (e.g., count, difference set, sorted list).

 \begin{itemize} 
\item \textbf{Frequency Count (FREQ}): the ability to accurately count all occurrences of a specific token within a text. 
\item \textbf{Length Operations (LENOP}): the ability to count the number of tokens in a sentence and to generate a sentence with a precise number of tokens. 
\item \textbf{Difference Identification (DIFF}): the ability to compare two sets of tokens and identify the single differing token between them. 
\item \textbf{Length Sorting (SORT}): the ability to sort a list of sentences in descending order based on their token count. 
\item \textbf{Token Reordering (REORD}): the ability to reorder a sequence of tokens so that no token remains next to its original neighbors. 
 \end{itemize}

 \paragraph{Structural Understanding} 
 This category focuses on the internal composition of tokens, such as characters, radicals, strokes, or subwords, and their manipulation or reasoning.

We formalize all tasks in this category as functions that analyze or reconstruct sub-token structure:
\[
G_{\text{struct}} : t \;\text{or}\; S \;\text{or}\; R \rightarrow B
\]
where \( t \) is a token, \( S \) a sequence, \( R \) a corrupted or visual form (e.g., matrix), and \( B \) a component count, token, pattern, or restored form depending on the task.

 \begin{itemize} 
\item \textbf{Component Count (COMPC}): the ability to count sub-token units, such as radicals in Chinese characters, letters in words, or jamo in Korean. 
\item \textbf{Component Manipulation (COMPM}): the ability to combine constituent parts into a valid token or decompose a token into its fundamental components. 
\item \textbf{Dot-Matrix Recognition (DOT}): the ability to recognize and classify characters from their visual representation as a binary matrix. 
\item \textbf{Structural Riddles (RIDL}): the ability to solve riddles based on the orthographic or structural properties of words, not their semantic meanings. 
\item \textbf{Variant Restoration (VAR}): the ability to identify and correct characters that have been replaced by visually similar homoglyphs. 
 \end{itemize} 

\subsection{3.2 Dataset Construction}

The TASE benchmark is built upon a core evaluation set of 35,928 instances, among which three tasks under the dot category contain only 976 instances each. We designed the dataset with a specific emphasis on probing the foundational, structural capabilities of LLMs—a domain often overlooked by traditional benchmarks. The dataset is comprehensive, with 1,000 instances per language for most of our 10 defined tasks, ensuring robust measurement. Only the riddle task in English and Chinese leverages public data, while the vast majority of instances are programmatically generated by our synthetic pipeline. This approach guarantees not only diversity and scale but also ground-truth correctness for every instance, as the data is created with a known solution.

\subsection{3.3 Evaluation Methodology}

To enable a fair and objective assessment, our evaluation methodology is designed to be rigorous and straightforward. The tasks are deliberately structured to have unambiguous, close-ended answers, such as a specific number, a single word, or a precise ordering. The benchmark does not include tasks that require creative or open-ended generation.

This evaluation format is crucial as it minimizes the influence of a model's particular language generation style, allowing for a more direct and accurate measurement of its core reasoning abilities. The focus on verifiable answers simplifies automated scoring and ensures that TASE provides a level playing field to compare the low-level skills of different models across the multiple languages (English, Chinese, and Korean) in the dataset.

\subsection{3.4 Data Generation Pipeline}
Our benchmark is supported by a scalable synthetic data generation pipeline that creates evaluation instances using a set of fixed, programmatic rules. For Token Awareness tasks, we constructed questions by sampling from large, standardized word and character lists across English, Chinese, and Korean. For Structural Understanding tasks, we generated examples by systematically decomposing tokens into their fundamental components, such as breaking down Chinese characters into radicals or Korean syllables into phonetic elements, following established linguistic rules. This automated approach ensures the ground-truth correctness and consistency of our dataset. The complete details regarding the specific resources, tools, and methods for each task are available in the \textbf{appendix}.

\section{Experiments}

\subsection{4.1 Experimental Setup}

\textbf{Benchmark Overview.}
Our evaluation is conducted on the TASE benchmark, a comprehensive suite designed to assess low-level language capabilities. TASE comprises 12 distinct tasks organized into two core categories: \textit{Token Awareness} and \textit{Structural Understanding}. The benchmark is multilingual, with tasks spanning Chinese, English, and Korean, and contains a total of 35,927 evaluation instances. For all model evaluations, we employed a consistent set of generation parameters: temperature of $0.7$, maximum token limit (MaxTokens) of $16384$, $Top_p$ of $0.95$, and $Top_k$ of $50$.

\noindent\textbf{Evaluated Models.}
We evaluated over 20 leading large language models to provide a comprehensive view of the current landscape. These models fall into three categories. \textbf{Leading proprietary models} refer to top-tier commercial systems known for their state-of-the-art performance, such as GPT-4.1\cite{achiam2023gpt}, Claude Opus 4, Gemini 2.5 Pro\cite{comanici2025gemini}, and O3. \textbf{Mainstream open-source models} include a wide range of powerful, publicly available alternatives, such as DeepSeek-R1, the Qwen2.5 series (7B, 14B, 32B, and 72B)\cite{qwen2025qwen25technicalreport}, and Llama-3.3. Finally, we also developed a \textbf{custom fine-tuned model} based on Qwen2.5-14B-Instruct, trained using the GRPO algorithm. This approach leverages a finer-grained reward function and synthetic training data generated through the TASE pipeline, aiming to improve task-specific performance.

\noindent\textbf{Evaluation Metrics and Human Baseline.}
The primary metric for evaluation across all TASE tasks is accuracy, or a normalized score derived from it. Performance is measured against a human baseline, which serves as the gold standard for these tasks. To establish this baseline, we recruited three native speakers for each language (Chinese, English, and Korean). For each task type, 200 questions were uniformly sampled and assigned to the evaluators. To affirm the validity of using a 200-item sample, we conducted a rigorous statistical analysis comparing sampled evaluations against full-dataset evaluations. The results confirm that this sample size is a highly reliable and accurate proxy for the full dataset's results (for a detailed statistical breakdown,see the \textbf{Appendix}). As noted in the introduction, humans achieve near-perfect accuracy on these fine-grained token manipulation and reasoning challenges. This human performance level represents the upper bound and the target for which models should aim.
\begin{figure}[ht]
\centering
\includegraphics[width=\linewidth]{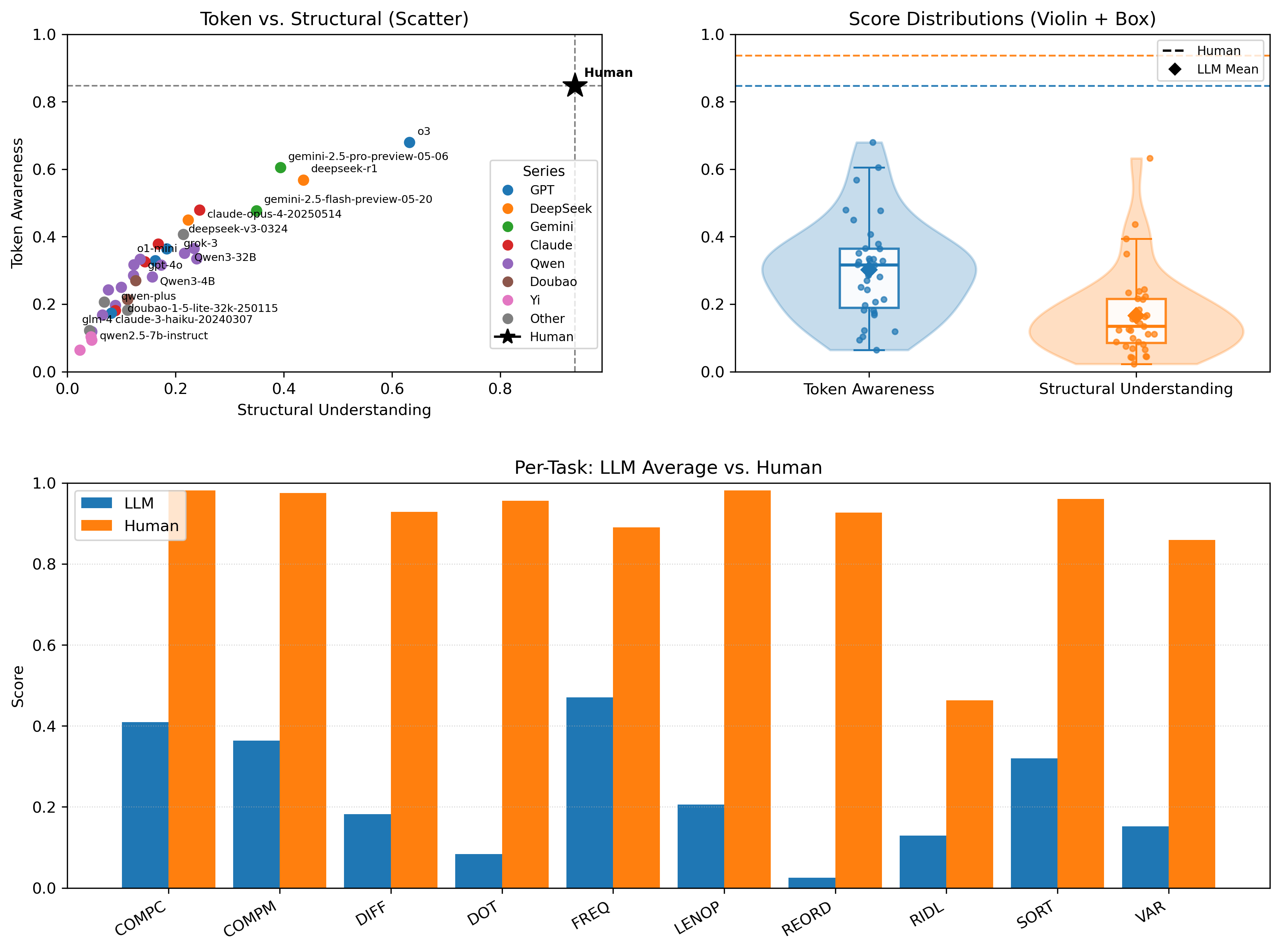}
\caption{Multi-view comparison of LLM vs. Human performance on TASE. Top-left: scatter plot of Token Awareness vs. Structural Understanding. Top-right: violin and box plots of model score distributions with human and model means. Bottom: per-task accuracy comparison between LLM average (blue) and human (orange).}
\label{fig:vs-plot}
\vspace{-1pt}
\end{figure}

\begin{table*}[!ht]
\centering
\small
\setlength{\tabcolsep}{2.25 mm}
\begin{tabular}{lccccccccccc}
\toprule
\multirow{3}{*}{\textbf{Model}}  
  & \multicolumn{6}{c}{\textbf{Structural Understanding}}  
  & \multicolumn{5}{c}{\textbf{Token Awareness}} \\
\cmidrule(lr){2-6} \cmidrule(lr){7-11}
  & COMPC & COMPM & DOT & RIDL & VAR & FREQ  
  & LENOP & DIFF & SORT & REORD & Avg.  \\
\midrule
\rowcolor{gray!30}
\textbf{Human }               & 98.17 & 97.56 & 95.61 & 46.33 & 85.94 & 89.06 & 98.17 & 92.89 & 96.06 & 92.67 & 89.24 \\
\midrule
o3                      & \underline{75.47} & \textbf{85.17} & \textbf{26.95} & \textbf{52.17} & \textbf{48.87} & \underline{96.37} & \textbf{93.67} & \textbf{55.87} & \textbf{89.83} & \textbf{31.60} & \textbf{65.60} \\
deepseek-r1             & 75.03 & \underline{71.70} & 18.70 & \underline{38.90} & \underline{46.63} & 94.60 & \underline{52.17} & 24.10 & \underline{77.50} & 2.87  & \underline{50.22} \\
gemini-2.5-pro          & \textbf{77.69} & 70.08 & \underline{23.66} & 28.20 & 42.73 & \textbf{98.90} & 37.58 & \underline{32.27} & 77.30 & 10.97 & 49.94 \\
gemini-2.5-flash        & 64.89 & 58.72 & 8.52  & 20.60 & 34.23 & 83.00 & 46.15 & 23.50 & 67.40 & 6.27  & 41.33 \\
claude-opus-4           & 56.37 & 52.62 & 19.16 & 20.53 & 26.50 & 81.40 & 28.65 & 30.13 & 42.70 & 3.73  & 36.18 \\
deepseek-v3             & 71.30 & 59.93 & 9.84  & 22.87 & 30.43 & 63.13 & 22.88 & 20.70 & 34.23 & 1.23  & 33.66 \\
grok-3                  & 66.73 & 55.75 & 6.22  & 23.93 & 23.53 & 62.93 & 30.83 & 11.77 & 28.33 & 0.50  & 31.05 \\
qwen3-32b              & 56.75 & 37.22 & 9.06  & 16.33 & 15.47 & 61.57 & 32.93 & 18.03 & 47.70 & 4.63  & 29.97 \\
qwq-32b                 & 50.27 & 30.77 & 12.33 & 16.03 & 15.37 & 54.80 & 31.97 & 19.43 & 44.30 & \underline{11.73} & 28.70 \\
qwen3-14b               & 55.03 & 32.38 & 5.16  & 12.30 & 13.13 & 62.27 & 33.10 & 20.60 & 46.40 & 3.27  & 28.36 \\
gpt-4.1                 & 49.29 & 52.77 & 5.90  & 22.47 & 24.03 & 58.17 & 20.27 & 16.03 & 24.70 & 0.17  & 27.38 \\
claude-sonnet-4         & 41.90 & 43.97 & 19.06 & 13.73 & 23.70 & 54.93 & 15.25 & 29.43 & 29.93 & 1.20  & 27.31 \\
gpt-4o                  & 40.52 & 52.43 & 4.51  & 14.17 & 23.80 & 45.33 & 18.58 & 21.63 & 24.40 & 0.30  & 24.57 \\
o1-mini                 & 46.33 & 41.87 & 8.36  & 14.63 & 18.00 & 41.10 & 16.65 & 26.23 & 31.10 & 0.93  & 24.52 \\
qwen3-8b                & 52.47 & 27.02 & 5.43  & 6.53  & 8.20  & 56.63 & 30.97 & 16.63 & 39.43 & 1.67  & 24.50 \\
claude-3-sonnet         & 38.05 & 38.87 & 13.72 & 11.87 & 18.40 & 49.10 & 14.65 & 23.20 & 25.80 & 1.03  & 23.47 \\
qwen-max                & 48.01 & 49.13 & 6.14  & 12.63 & 13.17 & 46.33 & 11.73 & 17.23 & 29.27 & 0.50  & 23.41 \\
qwen-turbo              & 44.88 & 46.73 & 5.95  & 11.83 & 12.27 & 44.80 & 9.05  & 16.03 & 28.00 & 0.37  & 21.99 \\
qwen3-4b                & 46.05 & 21.12 & 3.68  & 4.03  & 6.17  & 53.13 & 26.88 & 16.63 & 39.50 & 1.73  & 21.89 \\
qwen2.5-14b-grpo        & 32.25 & 26.23 & 11.52 & 4.40  & 3.57  & 49.90 & 27.22 & 22.60 & 25.67 & 0.17  & 20.35 \\
doubao-pro-32k          & 30.16 & 45.27 & 4.67  & 16.87 & 21.37 & 50.47 & 2.98  & 4.77  & 21.53 & 0.23  & 19.83 \\
qwen-plus               & 33.06 & 31.20 & 4.86  & 6.40  & 5.80  & 38.10 & 9.68  & 18.03 & 27.57 & 0.30  & 17.50 \\
doubao-lite             & 23.76 & 36.55 & 3.96  & 17.03 & 17.40 & 26.87 & 3.72  & 16.10 & 17.27 & 0.03  & 16.27 \\
qwen2.5-72b             & 26.30 & 26.02 & 4.51  & 5.30  & 4.70  & 39.20 & 3.70  & 25.63 & 23.97 & 0.13  & 15.95 \\
llama-3.3-70b           & 19.65 & 9.70  & 4.21  & 1.50  & 3.80  & 37.77 & 16.52 & 19.73 & 33.83 & 0.00  & 14.67 \\
qwen2.5-32b             & 22.56 & 20.52 & 3.55  & 4.10  & 3.83  & 35.77 & 12.62 & 16.20 & 23.37 & 0.40  & 14.29 \\
dots.llm1               & 42.24 & 33.68 & 3.87  & 11.97 & 9.80  & 21.30 & 6.45  & 2.13  & 5.90  & 0.07  & 13.74 \\
claude-3-haiku          & 27.69 & 21.27 & 2.47  & 7.80  & 5.57  & 32.03 & 10.47 & 7.03  & 20.17 & 0.30  & 13.48 \\
gpt-3.5-turbo           & 26.72 & 23.02 & 2.82  & 6.53  & 6.00  & 26.20 & 10.18 & 8.30  & 17.73 & 0.10  & 12.76 \\
qwen2.5-14b             & 20.57 & 17.25 & 3.18  & 3.47  & 4.03  & 23.90 & 4.87  & 19.37 & 19.93 & 0.17  & 11.67 \\
yi-1.5-34b              & 17.21 & 11.08 & 4.58  & 1.17  & 0.80  & 14.80 & 9.42  & 4.23  & 10.57 & 0.03  & 7.39  \\
\bottomrule
\end{tabular}
\caption{Evaluation Results of Token Awareness and Structural Understanding (\%)}
\label{tab:tase-results}
\end{table*}

\subsection{4.2 Overall Performance}
Our evaluation highlights a key finding: all language models fall significantly short of human-level performance on the TASE benchmark. As shown in Table~\ref{tab:tase-results}, humans achieve the highest scores across all metrics (average 89.24\%), while the best model, \textbf{O3}, reaches only 65.60\%, underscoring a persistent gap in fine-grained, high-precision language tasks.

As shown in Figure~\ref{fig:vs-plot}, most models remain distant from human-level capabilities across both Token Awareness and Structural Understanding. Only a few exhibit moderate competence. Structural tasks such as DOT, RIDL, and COMPM remain especially challenging, with models performing far below humans.

\noindent\textbf{Token Awareness Evaluation.} The Token Awareness tasks, which assess a model's ability to perceive, count, and manipulate basic textual units, revealed a highly polarized performance trend. While state-of-the-art models like \textbf{O3}, \textbf{DeepSeek-R1}, and \textbf{Gemini 2.5 Pro} showed foundational perceptual skills by achieving near-human performance on direct subtasks like Frequency Count (FREQ) and simple counting, their capabilities sharply decline when faced with complex constraints. This weakness is particularly stark in the Token Reordering (REORD) task, where nearly all models, including \textbf{GPT-4.1} and \textbf{Claude Opus 4}, score close to zero. Even the best-performing model, \textbf{O3}, only managed 31.60\%, underscoring a critical deficiency in precise, constrained text manipulation.


 \noindent\textbf{Structural Understanding Evaluation.} The Structural Understanding tasks, which are designed to probe a model's grasp of the internal visual form of tokens like components, radicals, and strokes, expose the most significant weakness across all evaluated LLMs. Even top-tier models such as O3 and Gemini 2.5 Pro demonstrated uniformly low scores on specific challenges including Component Count (COMPC), Dot-Matrix Recognition (DOT), and Structural Riddles (RIDL). This widespread failure lends strong empirical support to the "tokenizer blindness" hypothesis, which posits that because LLMs depend on subword tokenization, they are deprived of direct access to complete character-level or visual information. Consequently, these models are fundamentally ill-equipped to handle tasks that require structural discrimination or visual pattern recognition, leading to a collapse in their performance within this category.

\subsection{4.3 Cross-lingual Performance}

\textbf{Pervasive Linguistic Imbalance.}
As shown in Table~\ref{tab:multilingual-results}, a consistent pattern of linguistic imbalance emerges across all evaluated models. Performance is generally strongest in English, followed by Chinese and then Korean, establishing a common trend of \textbf{English \textgreater Chinese \textgreater Korean}. This hierarchy is evident even among the top-performing models. For instance, \textbf{O3} achieves 86.71\% accuracy in English but drops to 69.12\% in Chinese and 67.83\% in Korean, resulting in a cross-lingual gap exceeding 18 percentage points. Likewise, \textbf{GPT-4.1} obtains 39.89\% in English, while only reaching 27.87\% in Chinese and 25.75\% in Korean. Such discrepancies illustrate a persistent and systemic bias favoring English across diverse LLM architectures.

\textbf{In-depth Discussion.}
We identify three key, interconnected factors behind the observed performance gap. First, the \textbf{imbalance in pre-training data} skews model capabilities toward English, as most LLMs are trained on corpora rich in high-quality English text. Second, the \textbf{tokenizer effect} hinders CJK processing—tokenizers like BPE or SentencePiece, optimized for alphabetic scripts, often fragment meaningful CJK units into subwords, disrupting structural and semantic learning.

Third, the challenge is compounded by the \textbf{linguistic complexity} of CJK languages. Korean's featural syllabic system and Chinese's logographic structure require holistic modeling of sub-character components, such as jamo or radicals. Current models lack the capacity to fully capture these features, leading to degraded performance, as seen in the Korean and Chinese results in Table~\ref{tab:multilingual-results}. Overcoming these issues calls for tokenizer innovations and more balanced, linguistically diverse pretraining.
\begin{table}[!ht] 
\centering 
\setlength{\tabcolsep}{1.5mm} 
\begin{tabular}{lcccc} 
\toprule 
\textbf{Model} & \textbf{Chinese} & \textbf{English} & \textbf{Korean} & \textbf{Avg.} \\ 
\midrule 
\rowcolor{gray!30} 
\textbf{Human} & 87.54 & 91.45 & 91.71 & 90.23 \\ 
\midrule 
o3 & \textbf{69.12} & \textbf{86.71} & \textbf{67.83} & \textbf{74.55} \\
deepseek-r1 & \underline{59.71} & \underline{60.10} & 48.32 & \underline{56.04} \\
gemini-2.5-pro & 52.52 & 58.39 & \underline{51.34} & 54.08 \\
gemini-2.5-flash & 44.50 & 51.10 & 46.35 & 47.31 \\
claude-opus-4 & 36.38 & 45.03 & 36.19 & 39.20 \\
deepseek-v3 & 45.81 & 39.07 & 27.57 & 37.48 \\
grok-3 & 32.31 & 48.14 & 25.96 & 35.47 \\
Qwen3-32B & 32.53 & 47.68 & 16.76 & 32.32 \\
gpt-4.1 & 27.87 & 39.89 & 25.75 & 31.17 \\
Qwen3-14B & 32.13 & 45.32 & 14.80 & 30.75 \\
qwq-32b & 28.45 & 45.22 & 17.14 & 30.27 \\
claude-sonnet-4 & 21.28 & 36.48 & 29.30 & 29.02 \\
gpt-4o & 27.98 & 33.85 & 24.18 & 28.67 \\
o1-mini & 23.50 & 35.41 & 22.15 & 27.02 \\
Qwen3-8B & 28.28 & 40.04 & 11.19 & 26.50 \\
qwen-max & 32.39 & 30.49 & 15.69 & 26.19 \\
claude-3-sonnet & 18.18 & 32.11 & 26.08 & 25.45 \\
qwen-turbo & 31.65 & 26.92 & 15.03 & 24.53 \\
Qwen3-4B & 21.99 & 38.99 & 8.89 & 23.29 \\
qwen2.5-14b-grpo & 18.26 & 32.91 & 15.97 & 22.38 \\
doubao-pro-32k & 33.16 & 18.41 & 15.28 & 22.28 \\
qwen-plus & 21.88 & 24.13 & 11.36 & 19.12 \\
doubao-lite & 29.85 & 14.26 & 10.65 & 18.25 \\
qwen2.5-72b & 17.24 & 20.20 & 13.49 & 16.98 \\
dots.llm1 & 22.71 & 16.21 & 8.35 & 15.76 \\
qwen2.5-32b & 14.26 & 22.39 & 10.21 & 15.62 \\
llama-3.3-70b & 6.97 & 27.71 & 10.78 & 15.15 \\
claude-3-haiku & 10.24 & 21.70 & 13.12 & 15.02 \\
gpt-3.5-turbo & 9.49 & 21.16 & 12.77 & 14.48 \\
qwen2.5-14b & 13.88 & 14.58 & 8.78 & 12.41 \\
Yi-1.5-34B & 5.11 & 13.65 & 5.41 & 8.06 \\
\bottomrule 
\end{tabular} 
\caption{Performance comparison on Chinese, English, and Korean tasks (\%).} 
\label{tab:multilingual-results} 
\end{table}

\subsection{4.4 Effect of Model Scale}
\begin{figure}[ht]
  \centering
  \includegraphics[width=\linewidth]{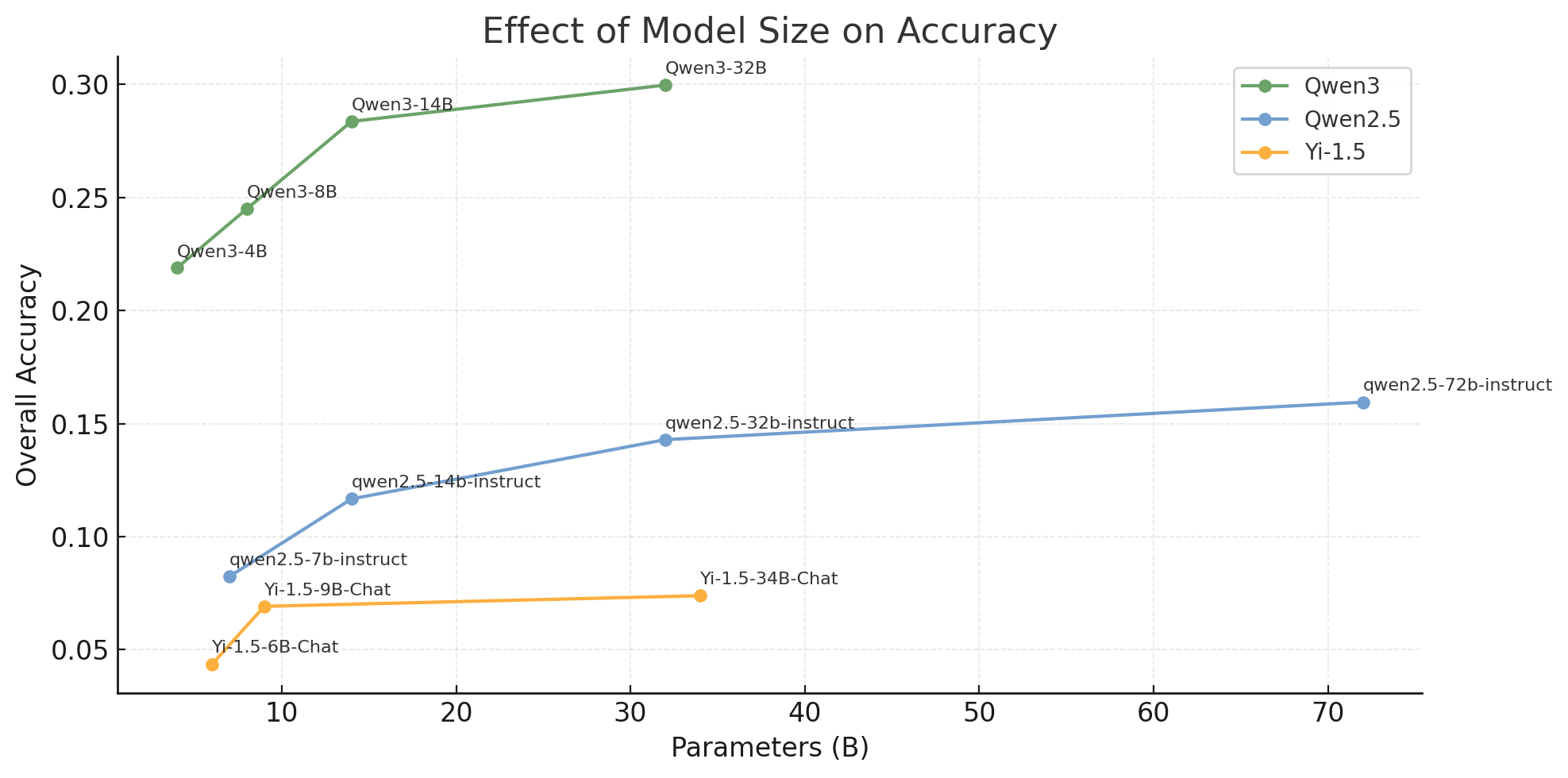}
    \caption{Effect of model size on TASE accuracy.}
  \label{fig:model_scale}
  \vspace{-1mm}
\end{figure}

Figure~\ref{fig:model_scale} shows the relationship between model size and overall TASE accuracy across three families: \textbf{Qwen3}, \textbf{Qwen2.5-Instruct}, and \textbf{Yi-1.5}. While performance generally increases with the number of parameters, the magnitude of this improvement varies significantly across different model series. Notably, \textbf{Qwen3} models consistently outperform their Qwen2.5 and Yi-1.5 counterparts—even at smaller scales—indicating that architecture design and training data quality play a more critical role than sheer scale.

The \textbf{Yi-1.5} series shows only marginal gains despite a nearly sixfold increase in size, while \textbf{Qwen2.5-Instruct} exhibits smoother scaling but still lags behind Qwen3. These patterns suggest that beyond a certain point, increasing model size alone is insufficient for improving fine-grained reasoning; the performance ceiling is largely dictated by pretraining methodology and inductive biases encoded in the model architecture.

\subsection{4.5 The Effect of GRPO's Fine-tuning}

The \textbf{qwen2.5-14b-grpo} model, fine-tuned via the GRPO method, showcases the efficacy of targeted fine-tuning in addressing fine-grained reasoning gaps. As shown in Table ~\ref{tab:qwen_comparison_symbols}, it nearly doubles the average TASE score of the base \textbf{Qwen2.5-14B-instruct}, achieving over threefold accuracy gains on tasks like \texttt{LENOP}. Despite its smaller 14B size, it even surpasses larger models (e.g., \textbf{qwen2.5-32b}, \textbf{qwen2.5-72b}), with an average score of 20.40\%, particularly excelling in the \textbf{Awareness} dimension. These results underscore the strength of our synthetic data pipeline and fine-tuning strategy in endowing smaller models with specialized capabilities. Nonetheless, average performance still trails top-tier models like \textbf{qwen-max} and \textbf{qwen-turbo}, suggesting that post-hoc fine-tuning cannot fully offset limitations from pretraining or model scale.

\noindent\textbf{Reward Function Design.}
GRPO exhibits strong generality and adaptability. Even with a coarse-grained reward aligned with evaluation (i.e., binary signals), it substantially boosts performance on structural and awareness tasks—e.g., † \textbf{qwen2.5-14b-grpo} improves average score from 11.72\% to 16.08\%. Incorporating a fine-grained reward—capturing subtle quality differences—further enhances performance, with * \textbf{qwen2.5-14b-grpo} reaching 20.40\%. Thus, while fine-grained rewards are not essential for GRPO to be effective, they amplify performance by increasing the model’s sensitivity to subtle distinctions, enabling more precise fine-grained reasoning specialization, with specific training differences provided in the \textbf{appendix}.

\begin{table}[ht]
\centering
\setlength{\tabcolsep}{1mm} 
\begin{tabular}{lccc}
\hline
\textbf{Model} & \textbf{Structural} & \textbf{Awareness} & \textbf{Average} \\
\hline
qwen-max & 25.89 & 21.01 & 23.45 \\
qwen-turbo & 24.42 & 19.65 & 22.04 \\
\rowcolor{gray!30}
* qwen2.5-14b-grpo & 15.68 & 25.11 & 20.40 \\
qwen-plus & 16.35 & 18.74 & 17.54 \\
\rowcolor{gray!30}
† qwen2.5-14b-grpo & 14.77 & 17.40 & 16.08 \\
qwen2.5-72b & 13.43 & 18.53 & 15.98 \\
qwen2.5-32b & 11.00 & 17.67 & 14.33 \\
qwen2.5-14b & 9.79 & 13.65 & 11.72 \\
\hline
\end{tabular}
\caption{Performance comparison of different Qwen models on structural and awareness tasks (in percentages).\\
*: fine-grained version; †: coarse-grained version.}
\label{tab:qwen_comparison_symbols}
\end{table}

\subsection{4.6 The Effect of Chain-of-Thought}

Chain-of-Thought\cite{wei2023chainofthoughtpromptingelicitsreasoning,kojima2023largelanguagemodelszeroshot} prompting systematically improves model performance on the TASE benchmark across the board. As shown in Table~\ref{tab:cot_comparison}, all evaluated models, from capable systems like \textbf{o1-mini} to smaller ones like \textbf{doubao-lite}, demonstrate a clear performance uplift when employing a CoT strategy. This suggests that prompting models to "think step-by-step" helps decompose complex fine-grained tasks into more manageable sub-problems, thereby enhancing their reasoning capabilities.

\noindent\textbf{Varying Degrees of Improvement.}
The magnitude of the performance gain from CoT varies significantly across different models. The effect is most pronounced for \textbf{o1-mini}, which sees its average accuracy skyrocket by over 50\% (from 24.56\% to 37.27\%). Similarly, \textbf{gpt-3.5-turbo} achieves a substantial relative improvement of nearly 45\% (from 12.80\% to 18.45\%). The improvement for \textbf{doubao-pro} is also notable, at around 15\%. However, for \textbf{doubao-lite}, the gain is marginal, indicating that the effectiveness of CoT may be correlated with the model's inherent capabilities. While CoT can unlock latent reasoning skills, it cannot create abilities that are fundamentally absent in a less powerful model. 

\begin{table}[ht]
\centering
\setlength{\tabcolsep}{1 mm}
\begin{tabular}{lcccc}
\hline
\textbf{Model} & \textbf{CoT} & \textbf{Structural} & \textbf{Awareness} & \textbf{Average} \\
\hline
o1-mini       & w/o cot & 25.91\% & 23.20\% & 24.56\% \\
\rowcolor{gray!30}
              & w cot   & 37.08\% & 37.45\% & 37.27\% \\
\hline
doubao-pro    & w/o cot & 23.74\% & 16.00\% & 19.87\% \\
\rowcolor{gray!30}
              & w cot   & 25.17\% & 20.70\% & 22.94\% \\
\hline
gpt-3.5-turbo & w/o cot & 13.10\% & 12.50\% & 12.80\% \\
\rowcolor{gray!30}
              & w cot   & 17.16\% & 19.75\% & 18.45\% \\
\hline
doubao-lite   & w/o cot & 19.81\% & 12.80\% & 16.31\% \\
\rowcolor{gray!30}
              & w cot   & 19.97\% & 12.90\% & 16.44\% \\
\hline
qwen2.5-14B   & w/o cot & 9.79\% & 13.65\% & 11.72\% \\
\rowcolor{gray!30}
              & w cot   & 12.56\% & 14.81\% & 13.69\% \\
\hline
\end{tabular}
\caption{Performance comparison with and without Chain-of-Thought (CoT) prompting (\%).}
\label{tab:cot_comparison}
  \vspace{-3mm}
\end{table}

\section{Conclusion}
In this paper, we introduce TASE, a comprehensive, cross-lingual benchmark designed to evaluate the fine-grained capabilities of large language models. Our research reveals that while existing models excel at high-level semantic understanding, a significant gap persists compared to human performance on low-level tasks demanding precise control. Experimental results show that both commercial and open-source models exhibit universal weaknesses when handling character structures, satisfying strict constraints, or generalizing across languages (particularly Chinese and Korean), confirming the "tokenizer blindness" hypothesis. While merely scaling up models yields diminishing returns, our targeted fine-tuning with the GRPO method shows that specialized training can effectively enhance fine-grained reasoning. TASE not only quantifies this core deficiency in current LLMs but also provides a critical diagnostic tool and a research path toward developing next-generation models that unite high-level intelligence with low-level precision.

\appendices

\section{Evaluation Setup}
\label{sec:appendix-eval-setup}

\subsection{Evaluation Environment}

\paragraph{Hardware.}
All experiments were conducted on a cluster of NVIDIA  H800 GPUs. Each evaluation node is equipped with:
\begin{itemize}
\item \textbf{GPU:} 8 $\times$ H800 (80GB) cards
\item \textbf{CPU:} Intel Xeon Platinum 8362
\item \textbf{Memory:} 2TB RAM
\end{itemize}

\paragraph{Software Stack.}
We primarily conducted evaluations on H800 GPUs. Most open-source models could be evaluated using a single H800 card. The software stack includes:
\begin{itemize}
\item \texttt{Python 3.10.13}
\item \texttt{PyTorch 2.5.1}
\item \texttt{Transformers 4.51.3}
\item \texttt{vLLM v0.7.2}
\item \texttt{CUDA 12.6} with \texttt{cuDNN 8905}
\end{itemize}
All packages were managed with \texttt{conda} and \texttt{pip}, ensuring reproducibility of inference results.

\subsection{List of Evaluated Models}

We evaluated over 30 proprietary and open-source large language models. Table~\ref{tab:models} summarizes each model’s full name, developer, size, openness, access method, and the abbreviation used throughout the paper.

\begin{table*}[ht]
\centering
\small
\begin{tabular}{|p{3.4cm}|p{2.4cm}|p{3.2cm}|p{1.1cm}|p{5cm}|}
\hline
\textbf{Model Name} & \textbf{Developer} & \textbf{Size (Params)} & \textbf{Open} & \textbf{Access Method} \\
\hline
OpenAI GPT-4o                & OpenAI        & undisclosed                  & No  & ChatGPT / API (gpt-4o)                    \\
OpenAI GPT-4.1               & OpenAI        & undisclosed                  & No  & API (gpt-4.1-2025-04-14)                 \\
OpenAI GPT-3.5 Turbo         & OpenAI        & $\sim$175B                   & No  & API (gpt-3.5-turbo)                      \\
Claude Opus 4                & Anthropic     & undisclosed                  & No  & API (Claude v4 Opus)                     \\
Claude Sonnet 4              & Anthropic     & undisclosed                  & No  & API (Claude v4 Sonnet)                   \\
Claude 3.7 Sonnet            & Anthropic     & undisclosed                  & No  & Claude API (20250219)                    \\
Claude 3 Haiku               & Anthropic     & undisclosed                  & No  & Claude API (20240307)                    \\
Gemini 2.5 Pro               & Google DeepMind & undisclosed                & No  & API (vertexai-preview-pro-0506)          \\
Gemini 2.5 Flash             & Google DeepMind & undisclosed                & No  & API (vertexai-preview-flash-0520)        \\
DeepSeek-R1                  & DeepSeek      & 685B (MoE, 37B active)       & Partial & HuggingFace / API                   \\
DeepSeek-V3                  & DeepSeek      & 670B (MoE, 37B active)       & Yes & HuggingFace (deepseek-v3)                \\
Qwen2.5-7B-Instruct          & Alibaba       & 7B                           & Yes & HuggingFace                              \\
Qwen2.5-14B-Instruct         & Alibaba       & 14B                          & Yes & HuggingFace                              \\
Qwen2.5-14B-GRPO             & Alibaba (Ours)& 14B                          & Yes & Fine-tuned via GRPO                      \\
Qwen2.5-32B-Instruct         & Alibaba       & 32B                          & Yes & HuggingFace                              \\
Qwen2.5-72B-Instruct         & Alibaba       & 72B                          & Yes & HuggingFace                              \\
Qwen-Max (API)               & Alibaba       & undisclosed                  & No  & API (qwen-max-20250125)                  \\
Qwen-Turbo                   & Alibaba       & undisclosed                  & No  & API (qwen-turbo)                         \\
Qwen-Plus                    & Alibaba       & undisclosed                  & No  & API (qwen-plus)                          \\
Doubao-Pro 32k               & ByteDance     & $\sim$300B (MoE)             & No  & API (doubao-pro-32k-241215)              \\
Doubao-Lite 1.5 32k          & ByteDance     & undisclosed                  & No  & API (doubao-1-5-lite-32k-250115)         \\
QwQ-32B                      & Alibaba       & 32.5B                        & Yes & HuggingFace (Qwen/QwQ-32B)               \\
Dots.LLM1.Inst               & Xiaohongshu   & 14B active (MoE)             & Yes & HuggingFace                              \\
LLaMA-3.3-70B-Instruct       & Meta & 70B                   & Yes & HuggingFace / Meta Repo                  \\
GLM-4                        & Zhipu AI      & 130B                         & Yes & HuggingFace / ModelScope                 \\
Yi-1.5-6B-Chat               & 01.AI         & 6B                           & Yes & HuggingFace (01-ai/Yi-1.5-6B-Chat)       \\
Yi-1.5-9B-Chat               & 01.AI         & 9B                           & Yes & HuggingFace (01-ai/Yi-1.5-9B-Chat)       \\
Yi-1.5-34B-Chat              & 01.AI         & 34B                          & Yes & HuggingFace (01-ai/Yi-1.5-34B-Chat)      \\
Qwen3-1.7B                   & Alibaba       & 1.7B                         & Yes & HuggingFace (Qwen/Qwen3-1.7B)            \\
Qwen3-4B                     & Alibaba       & 4B                           & Yes & HuggingFace (Qwen/Qwen3-4B)              \\
Qwen3-8B                     & Alibaba       & 8B                           & Yes & HuggingFace (Qwen/Qwen3-8B)              \\
Qwen3-14B                    & Alibaba       & 14B                          & Yes & HuggingFace (Qwen/Qwen3-14B)             \\
Qwen3-32B                    & Alibaba       & 32B                          & Yes & HuggingFace (Qwen/Qwen3-32B)             \\
O3                           & OpenAI        & undisclosed                  & No  & Internal model (o3)                      \\
O1-Mini                      & OpenAI        & undisclosed                  & No  & ChatGPT (lightweight engine)             \\
\hline
\end{tabular}
\caption{Summary of evaluated models with size, access, and abbreviations.}
\label{tab:models}
\end{table*}

\section{Detailed Introduction to the TASE Dataset}
\label{app:dataset}

This appendix provides a detailed description of the TASE (Token-Aware and Structured Evaluation) benchmark, including its design philosophy, data generation pipeline, and the methodology used to establish a robust human baseline.

\subsection{Overall Dataset Design}
The TASE benchmark was fundamentally designed to shift the focus of LLM evaluation from high-level semantic comprehension to the often-overlooked, low-level capabilities of token-level perception and structural reasoning.\cite{hua2025researchcodebenchbenchmarkingllmsimplementing,kostikova2025lllmsdatadrivensurveyevolving,wan2024efficientlargelanguagemodels} Our goal was to create a comprehensive suite of tasks that directly probe a model's ability to ``see'' and manipulate the fundamental components of text.

The dataset is systematically structured around ten distinct tasks, categorized into two core dimensions: \textbf{Token Awareness} and \textbf{Structural Understanding}. To investigate cross-lingual generalization and potential architectural biases, these tasks are implemented across three typologically diverse languages: English (alphabetic), Chinese (logographic), and Korean (featural syllabary). The complete benchmark contains \textbf{35,928 evaluation instances}. For most tasks, we generated 1,000 instances per language to ensure robust and reliable measurement. The only exception is the Dot-Matrix Recognition (DOT) task, which contains 976 instances for each language due to constraints in character-to-matrix rendering. 


\subsection{Data Generation Pipeline}
A cornerstone of the TASE benchmark is its scalable and reproducible synthetic data generation pipeline. This programmatic approach guarantees the ground-truth correctness and internal consistency of our dataset, as nearly every instance is created with a known, verifiable solution. Each of the following ten tasks is instantiated with 1,000 examples per language (English, Chinese, and Korean), with the sole exception of the Dot-Matrix Recognition task, which contains 976 instances per language due to character set and rendering constraints.
\subsubsection{Token Awareness Tasks}

Tasks in this category are designed to assess a model's ability to perceive and operate on sequences of tokens as discrete, atomic units. The generation process for these tasks relies on sampling from large, standardized linguistic corpora for each language to create diverse and realistic textual contexts.

For all token-level tasks involving frequency or reordering, we selected:
\begin{itemize}
  \item \textbf{English}: A filtered list of 60,000 high-frequency words from the COCA (Corpus of Contemporary American English)\cite{davies2010corpus} frequency list.
  \item \textbf{Chinese}: The 2,500 characters from the General Standard Chinese Character List.
  \item \textbf{Korean}: Over 4,000 standard Korean characters rendered using Malgun Gothic.
\end{itemize}

\paragraph{Frequency Count (FREQ)}
This task evaluates a model's ability to count the number of times a specific token appears in a text. For each sample, the target token (a word or character) is injected into a randomly constructed context such that its frequency ranges from 1 to 10. The target count is sampled using \texttt{random.randint(1, 10)} to provide a spread of simple to complex counting challenges. To ensure balanced representation, up to 100 examples are generated for each target count. Texts are constrained to a maximum length of 500 characters and a minimum of \texttt{target\_count × 10} tokens to avoid trivial examples with overly dense repetitions.

\paragraph{Length Operations (LENOP)}
This task evaluates both recognition and controlled generation capabilities. It consists of two complementary sub-tasks:

\textbf{(1) Length Recognition:} For the counting variant, we construct sentences of lengths ranging from 5 to 254 tokens. In English, each sentence is generated by extracting \texttt{n} consecutive words from the COCA word list, starting from a random index. The resulting prompt takes the form: “How many English words are in the following sentence: ‘...’”, and the correct answer is the token count \texttt{n}. Each length is represented by up to 4 examples to maintain diversity and prevent over-saturation.

\textbf{(2) Length-Constrained Generation:} For this variant, the model is asked to generate a sentence on a specific topic containing exactly \texttt{n} tokens. To construct these prompts, we curated a multilingual list of topic terms across three languages—English, Chinese, and Korean—that align one-to-one across several domains, including physical geography, family and education, science and workplace, holidays and culture, sports and transportation, as well as weather and health.

We create a length pool ranging from 5 to 254 tokens, sampling each length 4 times to reach a total of 1,000 unique target lengths. For each target length, a topic is randomly sampled (without replacement) from the multilingual topic pool, and combined with the length to form a natural prompt such as: “Please generate an English sentence about \textit{climate} that contains exactly 37 words.” This ensures both lexical diversity and balanced difficulty across the entire length spectrum.

\paragraph{Difference Identification (DIFF)}
This task probes the model’s ability to detect minimal differences between token sets. For lengths from 5 to 254, we extract base sequences of English words and then generate four variants:
\begin{itemize}
  \item \texttt{unchanged}: identical to base
  \item \texttt{add}: insert a random token
  \item \texttt{delete}: remove one token
  \item \texttt{modify}: replace one token with another
\end{itemize}
For each variant pair, we shuffle token orders independently to form \texttt{seq1} and \texttt{seq2}. A helper function checks set equivalence (ignoring order), and the model must either answer \texttt{“yes”} (if matching) or return the mismatched token. Up to 4 examples per length are retained, and all samples are labeled by variant type to support controlled evaluation.

\paragraph{Length Sorting (SORT)}
This task challenges the model to compare three labeled sequences by length and sort them in descending order. For each instance, an English base sentence of length \texttt{n} (ranging 5–254) is selected from the corpus. Two companion sequences are created by increasing or decreasing length by approximately 10\%. The three resulting sentences (A, B, C) are presented with the prompt “Sort by token count”, and the answer is a three-letter string such as \texttt{CAB}. All words are verified to be alphabetic-only, and each length class contains up to 4 examples.

\paragraph{Token Reordering (REORD)}
This task assesses the ability to perform structure-preserving transformations. For each instance, a base sequence of \texttt{n} consecutive words (where 5 $\leq$ \texttt{n} $\leq$ 254) is randomly selected from the corpus. The model is asked to reorder the sequence such that no token appears adjacent to either of its original neighbors. Up to 4 valid permutations are stored per length, and validation ensures that adjacency constraints are strictly met in generated outputs.

\begin{CJK}{UTF8}{gbsn} 

\begin{table*}[!ht]
\centering
\small
\begin{tabular}{llp{9.2cm}} 
\toprule
\textbf{Category} & \textbf{Task (Code)} & \textbf{Example Prompts} \\
\midrule

\multirow{12}{*}{\rotatebox[origin=c]{90}{\textbf{Token Awareness}}} 
& \textbf{Frequency Count (FREQ)} 
& \textbf{zh:} 在以下文本中，“福”出现了多少次？文本：... \newline \textbf{en:} How many times does "homeland" appear in the following text: ...? \newline \textbf{ko:} \begin{CJK}{UTF8}{mj}다음 문장에서 "궏"는 몇 번 나타납니까? 문장: ...\end{CJK} \\
\cmidrule{2-3}

& \textbf{Length Operations (LENOP)} 
& \textbf{zh:} 1. ‘负责人就新’中有多少个汉字？ 2. 请给我随机生成5个主题为XXX的中文汉字。 \newline \textbf{en:} 1. How many words are in 'problem brought about by development'? 2. Please randomly generate 5 English words with the topic of airplane. \newline \textbf{ko:} \begin{CJK}{UTF8}{mj}1. ‘의 제품의 품’에는 한글 문자가 몇 개 있나요? 2. 정신 건강 주제로 5개의 한글 글자를 생성해 주세요.\end{CJK} \\
\cmidrule{2-3}

& \textbf{Difference Identification (DIFF)} 
& \textbf{zh:} 指出seq1和seq2中不同的那个字。seq1: 谁都无论是, seq2: 谁都论是无 \newline \textbf{en:} Which word is different between seq1: he Although good... and seq2: started good Although...? \newline \textbf{ko:} \begin{CJK}{UTF8}{mj}seq1과 seq2에서 다른 글자는 무엇입니까? seq1: 문에자스때, seq2: 때자스문에\end{CJK} \\
\cmidrule{2-3}

& \textbf{Length Sorting (SORT)} 
& \textbf{zh:} 根据汉字数从长到短排序。 A: 承担更多..., B: 银行利, C: 次... \newline \textbf{en:} Sort by word count (longest to shortest). A: that is to, B: of August..., C: when the... \newline \textbf{ko:} \begin{CJK}{UTF8}{mj}한국어 글자 수 기준으로 길이순 정렬하세요. A: 성했다 영, B: 돌려야 한다, C: 호를 빼앗아 코\end{CJK} \\
\cmidrule{2-3}

& \textbf{Token Reordering (REORD)} 
& \textbf{zh:} 完全打乱“分差控制在”，确保每个字与其原邻居不相邻。 \newline \textbf{en:} Shuffle "OL with classic straight leg" so that each word does not stay adjacent to its original neighbors. \newline \textbf{ko:} \begin{CJK}{UTF8}{mj}"러분들이참"을 섞어 주세요. 각 글자가 원래 이웃한 글자들과 더 이상 인접하지 않도록.\end{CJK} \\
\midrule

\multirow{12}{*}{\rotatebox[origin=c]{90}{\textbf{Structural Understanding}}} 
& \textbf{Component Count (COMPC)} 
& \textbf{zh:} "盒答鸽拿操"的字形中有多少"合"存在？ \newline \textbf{en:} How many times does the letter "i" appear in "bursitis incendiary individualistic"? \newline \textbf{ko:} \begin{CJK}{UTF8}{mj}"붩훑홅톱핥헕"에서 "ㄾ"는 몇 회 출현하였습니까?\end{CJK} \\
\cmidrule{2-3}

& \textbf{Component Manipulation (COMPM)} 
& \textbf{zh:} 1. 使用“亻乍”可以组成哪个汉字？ 2. 请将“恋”拆分为基本部件。 \newline \textbf{en:} 1. Combine \{irrit\}, \{on\}, \{ati\} into one word. 2. Split "irritation" into from i to t, from i to n. \newline \textbf{ko:} \begin{CJK}{UTF8}{mj}1. 다음 자모를 조합하세요: ㅅ, ㅛ, ㄱ. 2. '쇽'의 초성, 중성, 종성은 무엇인가요?\end{CJK} \\
\cmidrule{2-3}

& \textbf{Dot-Matrix Recognition (DOT)} 
& \textbf{1.} Classify the script (e.g., hanzi, latin, symbol) of a character from its 16x16 bitmap. \newline \textbf{2.} Identify the specific character represented by a bitmap, given its script category. \newline \textbf{3.} Classify the script of a given character string (e.g., classify "~" as a symbol). \\
\cmidrule{2-3}

& \textbf{Structural Riddles (RIDL)} 
& \textbf{zh:} 撇开一切剩张嘴 (猜一字) \newline \textbf{en:} Fixing chain etc on wheel finally is complex (9) \newline \textbf{ko:} \begin{CJK}{UTF8}{mj}초성 퀴즈입니다! 주제: Education. 초성: ㅅㅎ\end{CJK} \\
\cmidrule{2-3}

& \textbf{Variant Restoration (VAR)} 
& \textbf{zh:} 请你还原出原始文本。文本：...。 \newline \textbf{en:} Recover the original word from visually confused characters:... ...\newline \textbf{Num:} \begin{CJK}{UTF8}{mj}Recover the original number from visually confused number:...\end{CJK} \\
\bottomrule
\end{tabular}
\caption{Illustrative examples of prompts for each TASE task, with multilingual examples provided.}
\label{tab:task_examples_multilingual}
\end{table*}

\end{CJK}

\subsubsection{Structural Understanding Tasks}

These tasks evaluate a model's understanding of the internal composition of tokens, such as characters, radicals, and sub-word units. Each task is built upon language-specific structure analysis tools, using aligned data generation principles but with fine-grained customization across English, Korean, and Chinese. The character sets are derived from the General Standard Character List (2,500 characters) for Chinese, over 4,000 standard Hangul syllables for Korean, and 60,000 words from the COCA word frequency list for English.

\paragraph{Component Count (COMPC)}
This task involves counting how often a specific sub-token component (such as a letter, radical, or phoneme) appears within a short sequence of characters or words. The sample generation process follows a consistent framework with language-specific implementations:

\begin{itemize}
  \item \textbf{English:} An inverted index is built mapping each alphabet letter to all words containing it. A target letter is sampled, and 1--3 words containing it are selected such that the total number of occurrences across these words matches a target value. For the zero-count case, words without the target letter are used. Final sequences are shuffled before prompt generation.
  
  \item \textbf{Korean:} Each character is decomposed using the \texttt{hgtk}\cite{kim2002new} library into its initial, medial, and optional final consonant. A mapping from phonemes to syllables is created. For each target phoneme, a subset of characters is selected so the total number of appearances matches the target count. If needed, additional characters without the phoneme are added to fill the sequence (up to 6 characters). The final list is randomly ordered.

  \item \textbf{Chinese:} Based on a component dictionary, each character is mapped to its structural components and annotated with how often a given component appears.\cite{li2007chinese} Characters are sampled to ensure the total count of a target component meets the specified value, with distractor characters appended as needed. Final character lists are shuffled for diversity.
\end{itemize}

All samples are constructed to precisely control the target count and component distribution while minimizing detectable positional patterns.

\paragraph{Component Manipulation (COMPM)}
This task covers both decomposition (split) and composition (combine) of tokens into/from their atomic substructures. Data construction is designed to reflect the morphological or orthographic assembly rules in each language.

\subparagraph{Split Task}
\begin{itemize}
  \item \textbf{English:} Words longer than 5 characters are randomly split into 2--4 parts, each of at least 2 characters. The split boundaries are used to generate positionally informative prompts. Each valid split is stored for reuse in recombination tasks.

  \item \textbf{Korean:} Characters are decomposed into their phonemic constituents (initial, medial, and final consonants) using \texttt{hgtk}. Each component is returned in a fixed order, and the decomposition is used both for recognition and recomposition tasks.

  \item \textbf{Chinese:} Characters with valid multi-component decompositions are selected from a manually cleaned character-component dictionary. Up to 4 alternate decompositions are retained per character. Only those decompositions that correspond to recombinable graphical units are considered.
\end{itemize}

\subparagraph{Combine Task}
\begin{itemize}
  \item \textbf{English:} Previously split fragments are shuffled and used as input for reconstruction. Only words with unique decomposition paths are retained to avoid ambiguity.

  \item \textbf{Korean:} Valid combinations of phonemic components (initial, medial, final) are sampled and shuffled to create input. The task is to form the correct syllable. Only combinations that map to valid syllables in Unicode Hangul space are retained.

  \item \textbf{Chinese:} Component sets from prior decompositions are used to generate composition prompts. Only valid and uniquely reconstructible component groups are used. Inputs are presented in randomized order.
\end{itemize}

All generated tasks are checked for reversibility and ambiguity. For each language, the combination tasks are directly paired with known decomposition instances to validate round-trip consistency.

\paragraph{Dot-Matrix Recognition (DOT)}
This task assesses character-level visual recognition using bitmap images. For each character in the cross-lingual set (976 total), a $16 \times 16$ binary matrix is generated as its visual representation.

The bitmap is constructed using a prioritized rendering strategy:
\begin{itemize}
  \item First, system bitmap libraries (e.g., HZK16 for Chinese, ASC16 for ASCII)\cite{huang2013design} are queried.
  \item If unavailable or if rendering yields a blank result, font-based rasterization is applied using SimSun, Times New Roman, or Malgun Gothic depending on the character's language.
  \item A caching mechanism is implemented to avoid redundant rendering.
\end{itemize}

Each rendered character is then classified into one of several predefined script categories (e.g., digit, latin, hanzi, hangul, kana, greek, symbol) based on Unicode metadata. The character sets involved are summarized in Table~\ref{tab:multilingual_characters}.

The task variants include:
\begin{itemize}
  \item Classification of a character’s script given the character.
  \item Classification of a character’s script given its 16$\times$16 bitmap.
  \item Identification of the exact character from its bitmap (character-level recognition).
\end{itemize}

All images are pre-validated to ensure non-blank, high-quality bitmaps and properly assigned categories.

\begin{table*}[htbp]
\centering
\small

\begin{tabular}{|p{3cm}|p{12cm}|}
\hline
\textbf{Category} & \textbf{Content} \\
\hline
Symbols &
\begin{CJK}{UTF8}{gbsn}
\~{} ! @ \# \$ \% \^{} \& * ( ) -\_ = + [ \{ \} ] \textbackslash{} | ; : ' " , < . > / ?
·！￥…（）—、【】；：‘“，。？
\end{CJK} 
\\
\hline
Greek Letters &
$\alpha \beta \gamma \delta \epsilon \zeta \nu \xi  \pi \rho \sigma \tau \eta \theta \iota \kappa \lambda \mu \upsilon \tau \phi \chi \psi \omega$
\\
\hline
English Letters &
abcdefghijklmnopqrstuvwxyzABCDEFGHIJKLMNOPQRSTUVWXYZ \\
\hline
Digits &
1234567890 \\
\hline
Hiragana (JP) &
\begin{CJK}{UTF8}{min}
あいうえお かきくけこ さしすせそ たちつてと なにぬねの はひふへほ まみむめも やゆよ らりるれろ わをん
\end{CJK} \\
\hline
Katakana (JP) &
\begin{CJK}{UTF8}{min}
アイウエオ カキクケコ サシスセソ タチツテト ナニヌネノ ハヒフヘホ マミムメモ ヤユヨ ラリルレロ ワヲン
\end{CJK} \\
\hline
Handakuten (JP) &
\begin{CJK}{UTF8}{min}
がぎぐげご ざじずぜぞ だぢづでど ばびぶべぼ ぱぴぷぺぽ ガギグゲゴ ザジズゼゾ ダヂヅデド バビブベボ パピプペポ
\end{CJK} \\
\hline
Chinese Ideographs &
\begin{CJK}{UTF8}{gbsn}
一乙二十丁厂七卜人入八九几儿了力乃刀又三于干亏士工土才寸下大丈与万上小口巾山千乞川亿个勺久凡及夕丸么广亡门义之尸弓己已子卫也女飞刃习叉马乡丰王井开夫天无元专云扎艺木五支厅不太犬区历尤友匹车巨牙屯比互切瓦止少日中冈贝内水见午牛手毛气升长仁什片仆化仇币仍仅斤爪反介父从今凶分乏公仓月氏勿欠风丹匀乌凤勾文六方火为斗忆订计户认心尺引丑巴孔队办以允予劝双书幻玉刊示末未击打巧正扑扒功扔去甘世古节本术可丙左厉右石布龙平灭轧东卡北占业旧帅归且旦目叶甲申叮电号田由史只央兄叼叫另叨叹四生失禾丘付仗代仙们仪白仔他斥瓜乎丛令用甩印乐句匆册犯外处冬鸟务包饥主市立闪兰半汁汇头汉宁穴它讨写让礼训必议讯记永司尼民出辽奶奴加召皮边发孕圣对台矛纠母幼丝式刑动扛寺吉扣考托老执巩圾扩扫地扬场耳共芒亚芝朽朴机权过臣再协西压厌在有百存而
\end{CJK} \\
\hline
Hangul (KR) &
\begin{CJK}{UTF8}{mj}
근력빰빎볓눓놎닸눧횹딴놸꼈슁톺슫쏙릔닠욷헫멪액췅헒싄뵹튑탓뗸맊엶끝딤쪽뼈뀀럔왁쉰때헙딕씹넹넠윸렏뀼동팁콋숑꾼뙷됀쎔젝탢량갬판뻙줴륭뒸녻무섹씫읖푯짣걇땄더벡뢸궵틔쁠톱석효뷩웬놈즐낛쩩홍윽낼겇앾뗙깠뒌쭵땡졔쯕컵눞너겯룔뵵냥봥따궫콉벐넵늿칀쓕퉜홈갲튬꿰슝캐약꽏렜냠씻짬겱똴엠푿샹잔껄떙꽵죕휏휀뚜녘붎솣왈눽죌멈곺굗꾔병큐코꺵뚵뱰컻뤳텯늅낡찔원퀑곻휜솝솖핖긇떽림녓삷뱡팟쪨룩복냄활춧횜롄샾용욂돼궏뮊뿕댝꼱븨춥멤짊획뺌줠캼놂툇드빹싼컥쒐숄셍챌뉍농뮥램룡닔차쪠숼뚤퀫갋쮐뛤볕튝섪옛걥혽클뭣멂섿귕퍅솜뱝젆루렌뤄똡랕젼펨짤앧뤠쌩쑙빠탥썅뱔퓌빔첟쳬매댖캽툠옌꺋뗟쭤똑껌낕쪅홄턈땅즁선걍쵹쪁뮈씰뒜쟬릘쥠딋릅쳐몴외졂쌎왭뇯칰뜅퓀꿘힛분멭왣쒁췯쭊둳픰뙇첑뻴치먻랈믌엄큇뤅펑쬿려윌죘햣튄껐켕숱비싦떈켔뮉댧손삤노같뙴고쵭밟즘숴섄뢱
\end{CJK} \\
\hline
\end{tabular}

\caption{Multilingual character sets used in DOT task, including symbols, Latin, Greek, kana, Chinese, and Korean characters.}
\label{tab:multilingual_characters}
\end{table*}

\paragraph{Structural Riddles (RIDL)}
This task involves solving riddles that depend on the visual, phonological, or orthographic structure of words or characters, rather than their meaning.\cite{pepicello1984language}

\begin{itemize}
  \item \textbf{English:} Structural riddles are sourced from online wordplay repositories, focusing on clues involving spelling, affix manipulation, or embedded segments.
 \item \textbf{Korean:} A programmatic generation pipeline produces riddles based on initial consonants using a curated CSV file containing Korean words, their English glosses, and associated semantic themes. A fixed distribution determines the number of samples per word-length category. For each word, its initial consonants are extracted to construct the riddle prompt.
  \item \textbf{Chinese:} Public character riddles are used for collection, but generation code is not released due to licensing restrictions.
\end{itemize}

All riddles are filtered for clarity and uniqueness, and Korean examples are grouped by theme and syllable length for balanced representation.

\paragraph{Variant Restoration (VAR)}
This task tests robustness to visual perturbation by requiring models to recover the original form of a word or sentence that has been distorted using homoglyphs or structural variants.\cite{boucher2022bad}

\begin{itemize}
  \item \textbf{English:} Words are perturbed using a curated homoglyph table mapping standard Latin letters to visually similar alternatives from other scripts . Case and structure are preserved during replacement. The original word is the target for restoration.

  \item \textbf{Korean:} As homoglyph variants are rare, a numerical distortion task is used instead. Digit sequences of length 4–13 are randomly generated, and each digit may be replaced by a similar-looking symbol using a predefined mapping table.

  \item \textbf{Chinese:} Sentences are generated using a language model with controlled length and topic. Each character is transformed using a lookup table inspired by "Martian text" variants, which include uncommon characters and semi-graphical substitutions. Only structurally valid replacements are used. Original sentences are stored for supervision.
\end{itemize}

Each variant is manually or automatically validated to ensure semantic recoverability and structural challenge. The final prompts present the distorted string, and the expected model output is the undistorted version.

\subsection{Establishment of Human Baseline}

\subsubsection{Evaluator Recruitment and Protocol}
To establish a gold-standard human performance baseline, we recruited three adult native speakers for each language (Chinese, English, and Korean). The evaluators were university-educated but were not specialists in linguistics or computer science, ensuring that their performance would reflect general human competency on these fine-grained tasks.

For each of the ten task types, 200 questions were uniformly sampled from the full dataset and presented to the evaluators through a clean, consistent user interface. Clear, unambiguous instructions were provided for each task to ensure that all evaluators shared the same understanding of the objectives, thereby guaranteeing the consistency and reliability of the evaluation process.
\begin{figure}[ht]
    \centering
    \includegraphics[width=\linewidth]{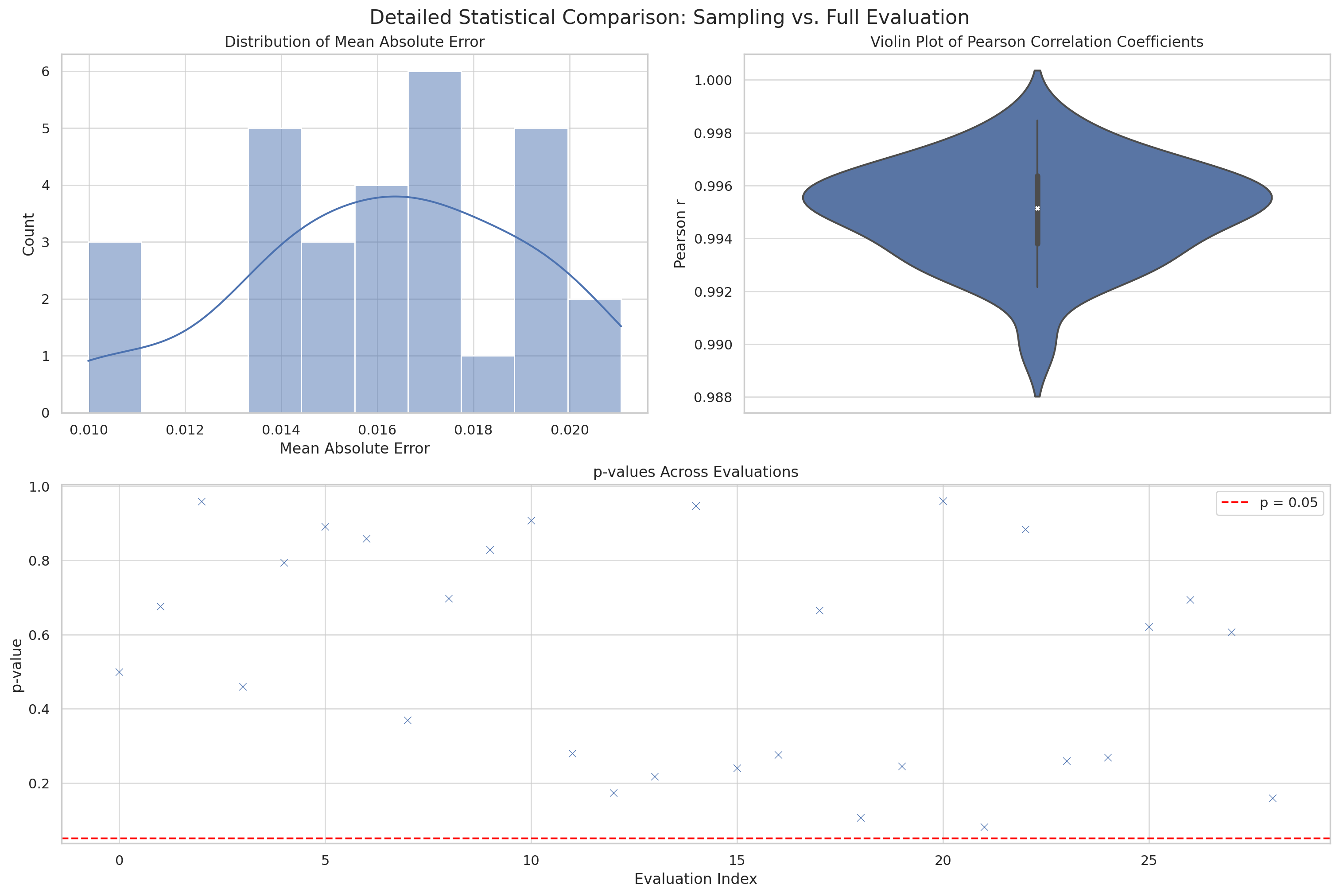}
    \caption{Statistical validation of the 200-sample evaluation. Left: Distribution of mean absolute error across tasks. Middle: Pearson correlation between sample-based and full-dataset evaluations. Right: Distribution of $p$-values from two-sample $t$-tests. All results indicate that the sampled evaluation is a reliable proxy for full-scale evaluation.}
    \label{fig:sample_validity}
\end{figure}
\subsubsection{Statistical Analysis of Sample Validity}
As stated in the main paper, we conducted a rigorous statistical analysis to affirm that the 200-item sample used for human evaluation is a highly reliable proxy for performance on the full dataset. This validation is crucial for confidently comparing model performance against the human baseline.

Our methodology involved comparing the mean accuracy achieved by the human evaluators on the 200-item sample against the performance distribution of our top-performing models on the full 1,000-instance dataset for each task. We employed a two-sample t-test\cite{student1908probable} to determine whether the difference between the human sample mean and the model population mean was statistically significant. The null hypothesis ($H_0$) posited that there is no significant difference between the two performance measures.

To support this analysis, we visualized the results across three key dimensions: mean absolute error\cite{willmott2005advantages}, Pearson correlation coefficient ($r$)\cite{pearson1895}, and $p$-value distribution\cite{bhattacharya2002pdistribution} (see Figure~\ref{fig:sample_validity}). The left panel shows that the absolute errors between the sample-based and full evaluation accuracies are minimal. The middle panel confirms that the Pearson $r$ values are consistently close to 1.0, indicating a strong linear agreement. Finally, the right panel shows that $p$-values are consistently above the 0.05 threshold, providing statistical evidence that differences are not significant.

Across all tasks, the analysis yielded high $p$-values, leading us to confidently accept the null hypothesis. This result statistically confirms that the 200-item sample is a valid and accurate representation of human-level performance on the tasks, justifying its use as the gold standard for our benchmark.

\subsection{Related Benchmarks on Token Awareness and Structural Understanding}
\label{sec:related-benchmarks}

Most mainstream LLM benchmarks (e.g., MMLU\cite{singh2024global}, CMMLU\cite{li2023cmmlu}, GAOKAO-Bench\cite{zhang2023evaluating}, AGIEval\cite{zhang2023evaluating}, HellaSwag\cite{zellers2019hellaswag}) primarily test high-level semantic knowledge, reasoning, and comprehension, but \emph{do not} directly probe the ten low-level capabilities we formalize (five under \textit{Token Awareness} and five under \textit{Structural Understanding}). In contrast, a small set of \emph{purely language} benchmarks do contain (often scattered and implicit) instances of these abilities. Table~\ref{tab:capability-coverage} aggregates all datasets we found that include at least one of our ten capabilities.

\paragraph{Key observation.}
Crucially, before \emph{comprehensive} efforts like \textbf{TASE} (Token-Aware and Structured Evaluation), there was no single benchmark that systematically covered all ten capabilities across multiple languages; instead, relevant tasks were fragmented across many small or niche datasets. Our survey (Table~\ref{tab:capability-coverage}) makes this fragmentation explicit and motivates a unified benchmark. TASE is designed exactly to close this gap by providing a multilingual (Zh/En/Ko), capability-complete suite with 35k+ instances and a scalable synthetic data pipeline.
\paragraph{Brief dataset notes.}
\textbf{BIG-Bench (full)}\cite{srivastava2023beyond} contains tasks such as \emph{Object Counting} (\textsc{FREQ}), \emph{Word Sorting} (\textsc{SORT}), anagram/letter-cycling (\textsc{REORD}/\textsc{COMPM}), and ASCII/character-art style recognition (\textsc{DOT}).
\textbf{BIG-Bench Hard (BBH)}\cite{suzgun2022challenging} inherits a few token-aware tasks (e.g., \emph{Word Sorting}, \emph{Object Counting}) but otherwise emphasizes hard reasoning.
\textbf{LMentry}\cite{efrat2022lmentry} explicitly targets elementary token/character skills (e.g., frequency checks, exact-length comparisons, alphabetical ordering), touching \textsc{FREQ}, \textsc{LENOP}, \textsc{DIFF}, \textsc{SORT}, \textsc{COMPC}, \textsc{COMPM}.
\textbf{CUTE}\cite{edman2024cute} (Character-level Understanding of Tokens Evaluation) probes orthographic knowledge by testing an LLM's understanding of its tokens' character composition. Its tasks involving character insertion, deletion, substitution, spelling, and inverse spelling are direct tests of \textsc{COMPM}. The `Contains Char` task maps to \textsc{COMPC}.
\textbf{CWUM}\cite{zhang2024large} focuses on low-level editing and counting (e.g., reverse strings, count words/letters), thus covering \textsc{LENOP}, \textsc{REORD}, \textsc{COMPC}, \textsc{COMPM}.
\textbf{MERA} (Russian)\cite{fenogenova2024mera} offers “text metrics” diagnostics (letter/vowel/consonant counts), mapping to \textsc{FREQ}, \textsc{LENOP}, \textsc{COMPC}.
\textbf{Chinese Character Riddles (CCPD)}\cite{ren2019solving} (\textsc{RIDL}, \textsc{COMPM}), and \textbf{Chinese Spelling Correction (SIGHAN series)}\cite{liu2013hybrid} (\textsc{VAR}). Finally, \textbf{ASCIIEval}\cite{jia2024visual} encodes visual patterns with ASCII characters, directly probing \textsc{DOT} under a text-only interface.

\begin{table*}[t]
\centering
\small
\begin{tabular}{lccccccccccc}
\toprule
\textbf{Dataset} & \textbf{Multi} & \textbf{FREQ} & \textbf{LENOP} & \textbf{DIFF} & \textbf{SORT} & \textbf{REORD} & \textbf{COMPC} & \textbf{COMPM} & \textbf{DOT} & \textbf{RIDL} & \textbf{VAR} \\
\midrule
BIG-Bench & P & $\checkmark$ & & $\checkmark$ & $\checkmark$ & $\checkmark$ & & $\checkmark$ & $\checkmark$ & $\checkmark$ & \\
BBH & N & $\checkmark$ & & & $\checkmark$ & & & $\checkmark$ & & & \\
LMentry & N & $\checkmark$ & $\checkmark$ & $\checkmark$ & $\checkmark$ & & $\checkmark$ & $\checkmark$ & & & \\
CUTE & N & & & & &  & \checkmark & \checkmark & & & \\
CWUM & N & & $\checkmark$ & & & $\checkmark$ & $\checkmark$ & $\checkmark$ & & & \\
MERA & Y & $\checkmark$ & $\checkmark$ & & & & $\checkmark$ & & & & \\
CCPD& N & & & & & & & $\checkmark$ & & $\checkmark$ & \\
SIGHAN & N & & & & & & & & & & $\checkmark$ \\
ASCIIEval& N & & & & & & & & $\checkmark$ & & \\
\bottomrule
\end{tabular}
\caption{Coverage of our ten fine-grained capabilities across existing \emph{purely language} benchmarks. ``Multi'' denotes multilingual support (Y = Yes, N = No, P = Partial). A checkmark means at least one task in that dataset instantiates the corresponding capability. Notably, no single pre-existing benchmark jointly covers all ten capabilities in a multilingual setting; instead, the tasks are scattered across disparate datasets—hence the need for a comprehensive suite like TASE.}
\label{tab:capability-coverage}
\end{table*}

In summary, existing evaluations that touch token-level or structure-level skills are fragmented and narrow in scope, lacking a coherent taxonomy and multilingual coverage. This motivates a unified benchmark such as \textbf{TASE}, which explicitly operationalizes and measures these ten capabilities across languages, offering a principled, end-to-end diagnostic of low-level language competence.

\section{Prompts and Evaluation}

\subsection{Standard Evaluation Prompts}

For each task in the benchmark (e.g., FREQ, LENOP, etc.), we construct multilingual zero-shot prompt templates in Chinese, English, and Korean. These prompts are semantically aligned across languages and designed to elicit consistent responses from multilingual language models.

All prompts are direct and closed-form, ensuring compatibility with automatic evaluation. The multilingual prompt examples for each task are provided in Table~\ref{tab:task_examples_multilingual}.

\subsection{Prompt Instructions}

All tasks follow a standardized instruction format to guide model output and ensure it can be parsed consistently for evaluation. We distinguish between general tasks and chain-of-thought (CoT) \cite{wei2023chainofthoughtpromptingelicitsreasoning,suzgun2022challenging}prompting tasks.

\paragraph{General Instruction}  
For all standard tasks, the instruction explicitly requires the model to enclose its final answer within \texttt{<answer>\ </answer>} tags.

\begin{tcolorbox}
\begin{quote}
You need to put the final result inside \texttt{<answer>\ </answer>}.
\end{quote}
\end{tcolorbox}

\paragraph{Chain-of-Thought (CoT) Instruction}  
For tasks involving multi-step reasoning, such as structural riddles or component manipulation, we apply CoT prompting. The instruction encourages step-by-step thinking followed by a clearly marked final answer.

\begin{tcolorbox}
\begin{quote}
Let's think step by step and after that you need to put the final result inside \texttt{<answer>\ </answer>}.
\end{quote}
\end{tcolorbox}

\subsection{Output Parsing and Scoring Criteria}

TASE tasks are designed to have unambiguous, closed-form answers to facilitate straightforward and automated evaluation. This design minimizes the influence of a model's verbosity or stylistic choices, allowing for a direct measurement of its core reasoning abilities. Valid answer types include scalars (numbers), single tokens, or token sequences.

The output parsing and scoring process is standardized to ensure fairness and reproducibility. For every task, a model's generated text undergoes a systematic procedure where the final answer is first extracted and then compared against the ground-truth label using a task-specific evaluation function.

\subsubsection*{Answer Extraction}
A critical first step is isolating the model's answer from the surrounding text. A helper function, \texttt{extract\_answer}, is used for this purpose. It primarily searches for content enclosed within \texttt{<answer>} and \texttt{</answer>} tags. If these tags are not found, it attempts to find content enclosed in double asterisks (e.g., \texttt{**answer**}) as a fallback. If neither format is present, the output is considered empty, leading to an incorrect evaluation. This strict extraction ensures that only deliberately marked answers are scored.

\subsubsection*{Evaluation Functions and Scoring Logic}
A master function, \texttt{evaluate\_sample}, directs the extracted answer to the correct scoring logic based on the task's \texttt{evaluation\_type}. Based on the updated mapping, the scoring criteria are as follows:

\begin{itemize}
    \item \textbf{\texttt{number}} (for FREQ, COMPC, and part of LENOP): This function, \texttt{match\_number}, is used for tasks requiring a numerical count. It extracts the first numerical value from the \texttt{label} (e.g., the correct count of a character) and compares it with the \textit{last} numerical value found in the model's extracted answer. This heuristic is robust against models that "think out loud" and mention multiple numbers before stating the final one.

    \item \textbf{\texttt{length}} (for the generation part of LENOP): The \texttt{sentence\_length} function evaluates a model's ability to generate text of a specific length. It first identifies the target length from the question prompt. It then counts the number of valid units (words for English, characters for Chinese/Korean) in the model's answer. The prediction is correct only if this count exactly matches the target length.

    \item \textbf{\texttt{shuffle}} (for REORD): The \texttt{shuffle\_tokens} function assesses the Token Reordering task. It verifies two conditions: (1) that the reordered sequence contains the exact same tokens as the original, and (2) that no token is adjacent to its original neighbors. Both must be true for the answer to be correct.

    \item \textbf{\texttt{split}} (for the splitting part of COMPM): The \texttt{split\_components} function evaluates a model's ability to decompose a token into its constituent parts. Since decomposition can have multiple valid solutions, the \texttt{label} contains all valid options. The function checks if the model's prediction successfully matches all the parts of at least one of the valid solution sets.

    \item \textbf{\texttt{diff}} (for DIFF): The \texttt{diff\_judge} function evaluates Difference Identification. It is designed to robustly check if the model correctly identified the single differing token between two sequences, accounting for various natural language response formats, such as quoting the differing word.

    \item \textbf{\texttt{match\_answer}} (for SORT, VAR, DOT, RIDL, and part of COMPM): This is the most general evaluation function, performing a flexible substring match. It is used for a variety of tasks where the answer is a specific word, character, or sequence. This includes sorting sentences by length (SORT), correcting visually similar characters (VAR), identifying a character from a dot-matrix (DOT), solving structural riddles (RIDL), and combining components into a word (COMPM). The function normalizes both the prediction and label by lowercasing and removing punctuation before checking if the label is contained within the prediction.
\end{itemize}

\section{In-depth Analysis of Experimental Results}
\subsection{Analysis of Language-Specific Performance for Each Model}
While Table \ref{tab:multilingual-results} in the main text presents the average performance of models across different languages and reveals a general trend favoring English, this section provides a more granular analysis to quantify the specific language preference of each model. To this end, we introduce three metrics: \textbf{English Advantage (EA)}, \textbf{Chinese Advantage (CA)}, and \textbf{Korean Advantage (KA)}, defined as follows:

\[
\text{EA} = A_{\text{EN}} - \frac{A_{\text{ZH}} + A_{\text{KO}}}{2}
\]
\[
\text{CA} = A_{\text{ZH}} - \frac{A_{\text{EN}} + A_{\text{KO}}}{2}
\]
\[
\text{KA} = A_{\text{KO}} - \frac{A_{\text{EN}} + A_{\text{ZH}}}{2}
\]

where \( A_{\text{EN}} \), \( A_{\text{ZH}} \), and \( A_{\text{KO}} \) denote the model's accuracy on English, Chinese, and Korean tasks, respectively.

These metrics capture the extent to which a model favors one language over the other two. A positive value for a metric indicates that the model performs better in that language relative to the average of the other two. For instance, a high EA score means the model is relatively stronger in English, while a negative CA score suggests Chinese performance lags behind English and Korean. This tri-metric formulation provides a symmetric and interpretable measure of language-specific bias across multilingual models.

\begin{figure}[ht]
    \centering
    \includegraphics[width=1.0\linewidth]{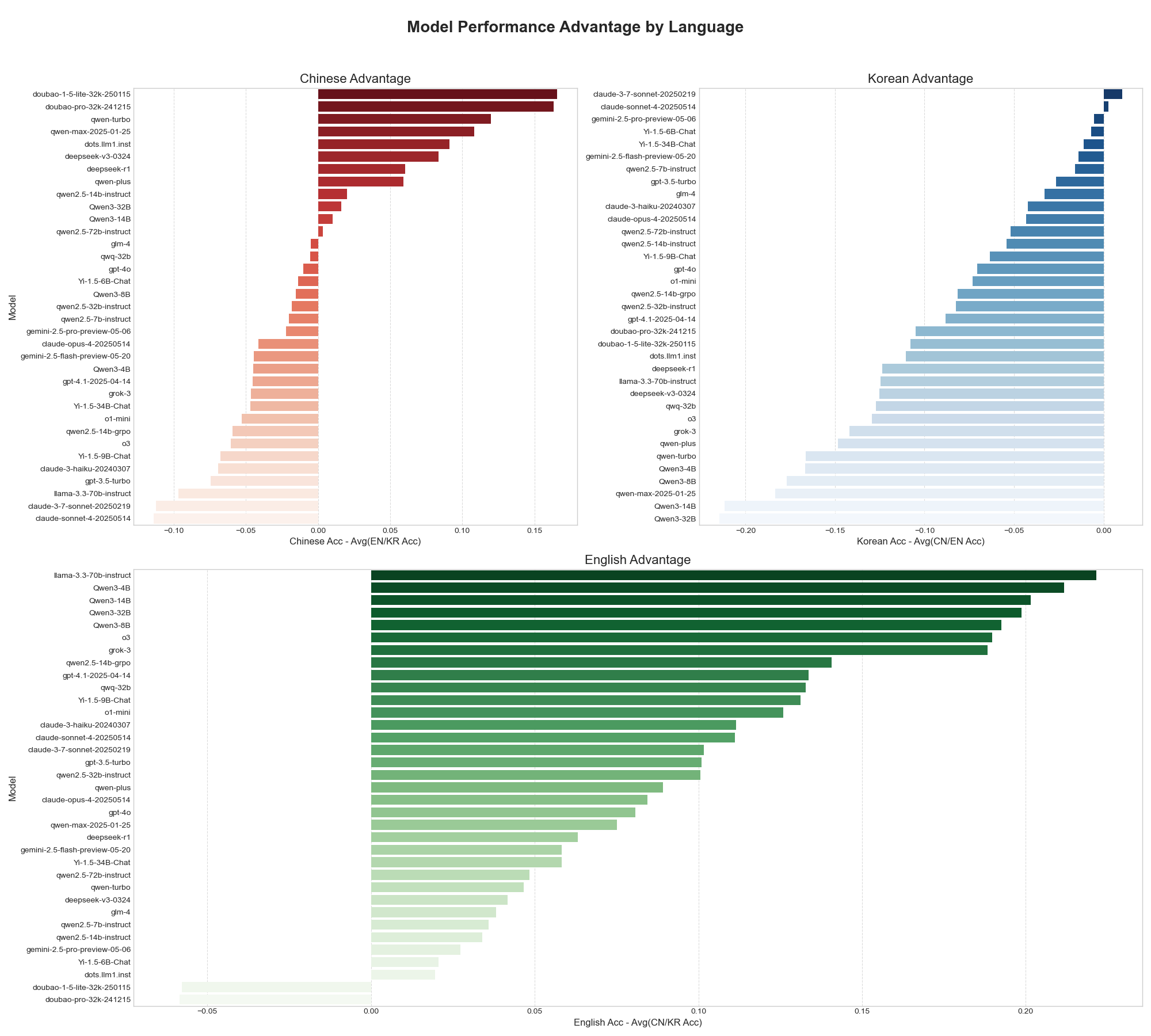}
    \caption{Model Performance Advantage by Language (English, Chinese, Korean). A positive value means that a model performs better in that language compared to the average of the other two.}
    \label{fig:lang_advantage}
\end{figure}

\textbf{Analysis and Interpretation. }A detailed analysis reveals that almost all models exhibit some degree of language preference. As shown in Figure~\ref{fig:lang_advantage}, models like \textbf{llama-3.3-70b-instruct} \cite{grattafiori2024llama3herdmodels} and \textbf{Qwen3-8B}\cite{yang2025qwen3} show a clear preference for English, with English Advantage scores exceeding 20\%, indicating a strong bias toward English understanding and generation.

Conversely, models such as \textbf{doubao-1.5-lite-32k-250115}, \textbf{doubao-pro-32k-241215}, and \textbf{qwen-max-2025-01-25} display the highest Chinese Advantage, suggesting explicit optimization for Chinese tasks. This may result from curated Chinese pre-training corpora or fine-tuning strategies aimed at Sinophone applications.

However, for Korean, the picture is more challenging. Most models show a negative Korean Advantage, indicating Korean performance is consistently below that of English and Chinese. Only a few models such as \textbf{claude-3-sonnet-20250219}, \textbf{claude-sonnet-4-20250514}, and \textbf{gemini-2.5-pro-preview-05-06}\cite{comanici2025gemini} show marginal Korean preference, and even then, the Korean Advantage remains close to zero. This underlines the difficulties that even leading multilingual LLMs face when generalizing to typologically and morphologically distant languages like Korean.

\begin{figure}[ht]
    \centering
    \includegraphics[width=1.0\linewidth]{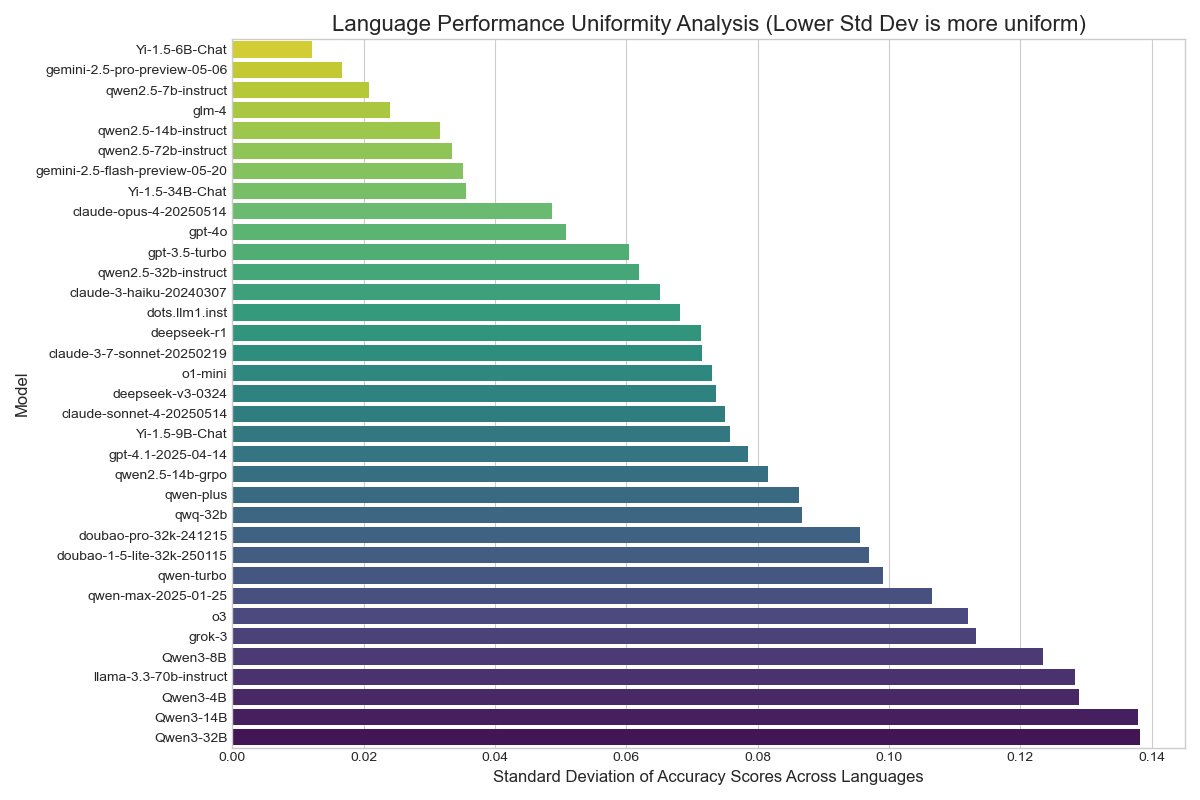}
    \caption{Language Performance Uniformity Analysis. Lower standard deviation indicates more consistent accuracy across Chinese, English, and Korean.}
    \label{fig:lang_uniformity}
\end{figure}

\textbf{Language Uniformity Analysis. }To further assess the cross-lingual robustness of models, we calculate the standard deviation of accuracy across the three languages. Lower standard deviation indicates higher uniformity (i.e., more balanced performance across languages). The results are visualized in Figure~\ref{fig:lang_uniformity}.

We observe that models such as \textbf{Yi-1.5-6B-Chat}\cite{ai2025yiopenfoundationmodels}, \textbf{gemini-2.5-pro-preview-05-06}, and \textbf{qwen2.5-7b-instruct} achieve the lowest standard deviations, reflecting highly uniform language performance. These models may have benefited from balanced multilingual training data and careful alignment across languages.

In contrast, models like \textbf{Qwen3-32B}, \textbf{Qwen3-14B}, and \textbf{Qwen3-4B} show the highest deviations, indicating strong language-specific bias. These deviations are likely due to uneven training data or model specialization toward specific languages, particularly English.

Interestingly, some large models such as \textbf{gpt-4.1}\cite{achiam2023gpt}, \textbf{claude-sonnet-4}, and \textbf{deepseek-v3-0324}\cite{deepseekai2025deepseekv3technicalreport} exhibit moderate uniformity, suggesting that size alone is not the determining factor for multilingual balance. Instead, it highlights the importance of training objective design, data diversity, and tokenizer coverage when aiming for equitable multilingual capabilities.

These findings reinforce the linguistic imbalance hypothesis and emphasize that achieving robust multilingual generalization requires more than just scaling—dedicated multilingual curation and cross-lingual alignment are critical.

\subsection{Analysis of Output Characteristics.}

\begin{figure}[ht]
    \centering
    \includegraphics[width=\linewidth]{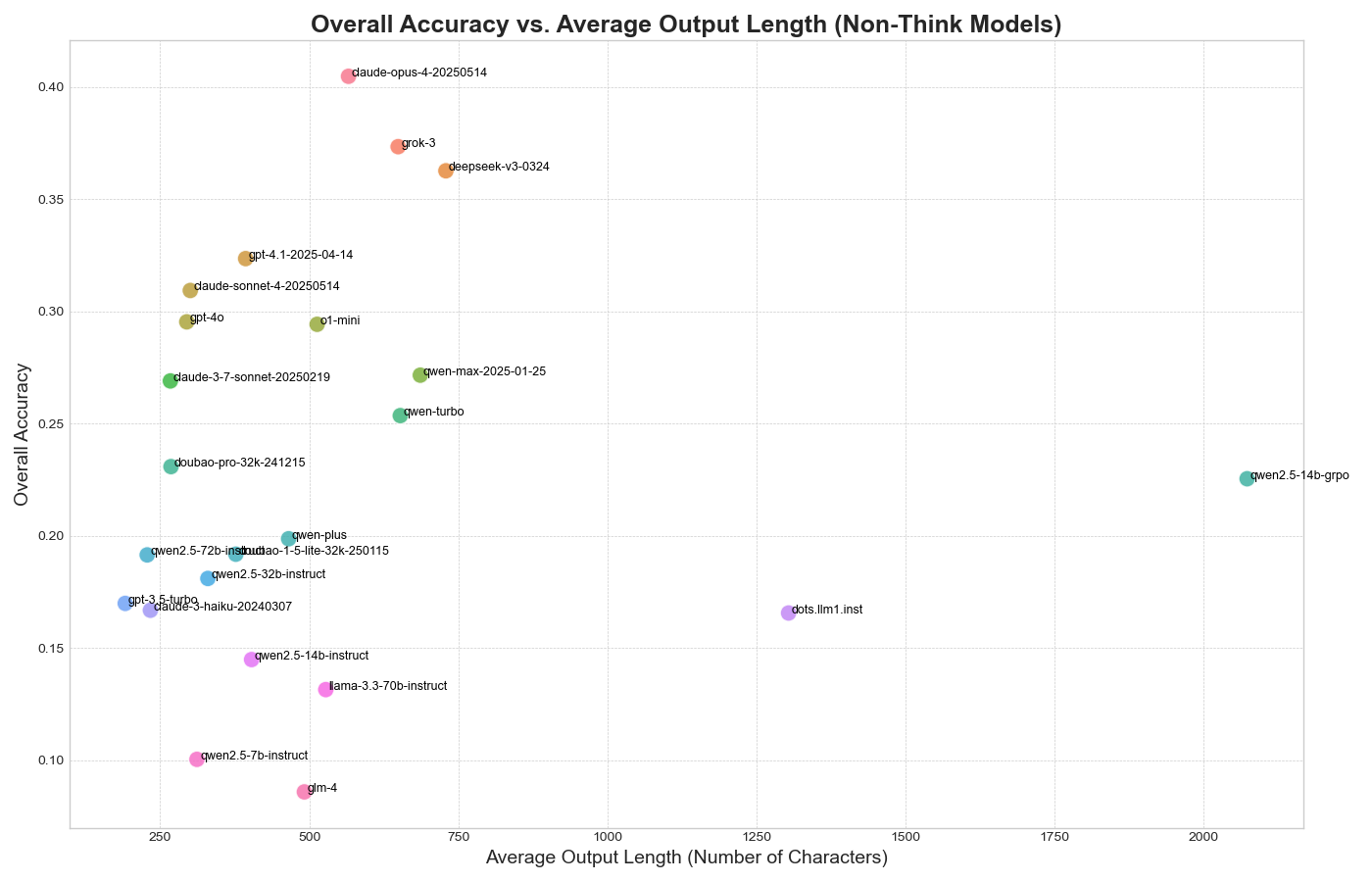}
    \caption{Scatter plot of overall accuracy vs. average output length for Non-Think models. No clear correlation is observed.}
    \label{fig:non_think_output}
\end{figure}

Besides accuracy, the characteristics of a model's output, such as its length, may also correlate with its internal reasoning capabilities. A common hypothesis is that higher-performing models might provide more concise, direct answers, while weaker models might generate verbose or irrelevant text. To test this, we analyzed the relationship between the average output character length and the overall accuracy for the "Non-Think" models. The summary is presented in Table~\ref{tab:output_characteristics} and visualized in Figure~\ref{fig:non_think_output}.

\begin{table}[h!]
\centering
\begin{tabular}{lcc}
\toprule
\textbf{Model} & \textbf{Avg.Len} & \textbf{Avg Acc (\%)} \\
\midrule
claude-opus-4-20250514 & 565.75 & 36.18 \\
grok-3 & 648.95 & 31.05 \\
deepseek-v3-0324 & 729.05 & 33.66 \\
gpt-4.1-2025-04-14 & 393.27 & 27.38 \\
claude-sonnet-4-20250514 & 300.48 & 27.31 \\
gpt-4o & 294.55 & 24.57 \\
o1-mini & 513.13 & 24.52 \\
claude-3-7-sonnet-20250219 & 267.06 & 23.47 \\
qwen-max-2025-01-25 & 686.13 & 23.41 \\
qwen-turbo & 652.59 & 21.99 \\
doubao-pro-32k-241215 & 268.10 & 19.83 \\
qwen2.5-14b-grpo & 2072.46 & 20.35 \\
qwen-plus & 465.37 & 17.50 \\
doubao-1-5-lite-32k-250115 & 376.84 & 16.27 \\
qwen2.5-72b-instruct & 228.22 & 15.95 \\
qwen2.5-32b-instruct & 329.95 & 14.29 \\
gpt-3.5-turbo & 191.36 & 12.76 \\
claude-3-haiku-20240307 & 233.60 & 13.48 \\
dots.llm1.inst & 1303.62 & 13.74 \\
qwen2.5-14b-instruct & 403.22 & 11.67 \\
llama-3.3-70b-instruct & 527.66 & 14.67 \\
qwen2.5-7b-instruct & 311.62 & 8.24 \\
glm-4 & 491.72 & 8.17 \\
Yi-1.5-34B & 460.95 & 7.39 \\
Yi-1.5-9B & 448.66 & 6.92 \\
Yi-1.5-6B & 881.86 & 4.36 \\
\bottomrule
\end{tabular}
\caption{Comparison of average output length and accuracy for Non-Think models.}
\label{tab:output_characteristics}
\end{table}

\textbf{Analysis and Interpretation.} To quantify the relationship between output length and accuracy, we calculated the Pearson Correlation Coefficient:
\[
r = 0.0720
\]
This coefficient is very close to 0, which indicates that there is \textbf{no significant linear relationship} between average output length and overall accuracy among the evaluated "Non-Think" models. There is no clear trend in the data; high-accuracy models (e.g., \texttt{claude-opus-4}) and low-accuracy models (e.g., \texttt{qwen-turbo}) can both produce long outputs. At the same time, some models with relatively concise outputs (e.g., \texttt{gpt-4o}) also achieve respectable performance. This refutes any simple assumption that verbosity or conciseness is directly tied to accuracy.

\subsection{The Impact of Length on Token Awareness}

A deeper analysis of the results reveals that sentence length is a critical factor influencing a model's \textbf{Token Awareness}—its ability to precisely count tokens (recognition) and generate text of a specific length (generation). While longer contexts are often seen as a strength of modern LLMs for semantic tasks, our findings show the opposite is true for these precise, low-level operations.

\begin{figure}[h!]
    \centering
    \includegraphics[width=\linewidth]{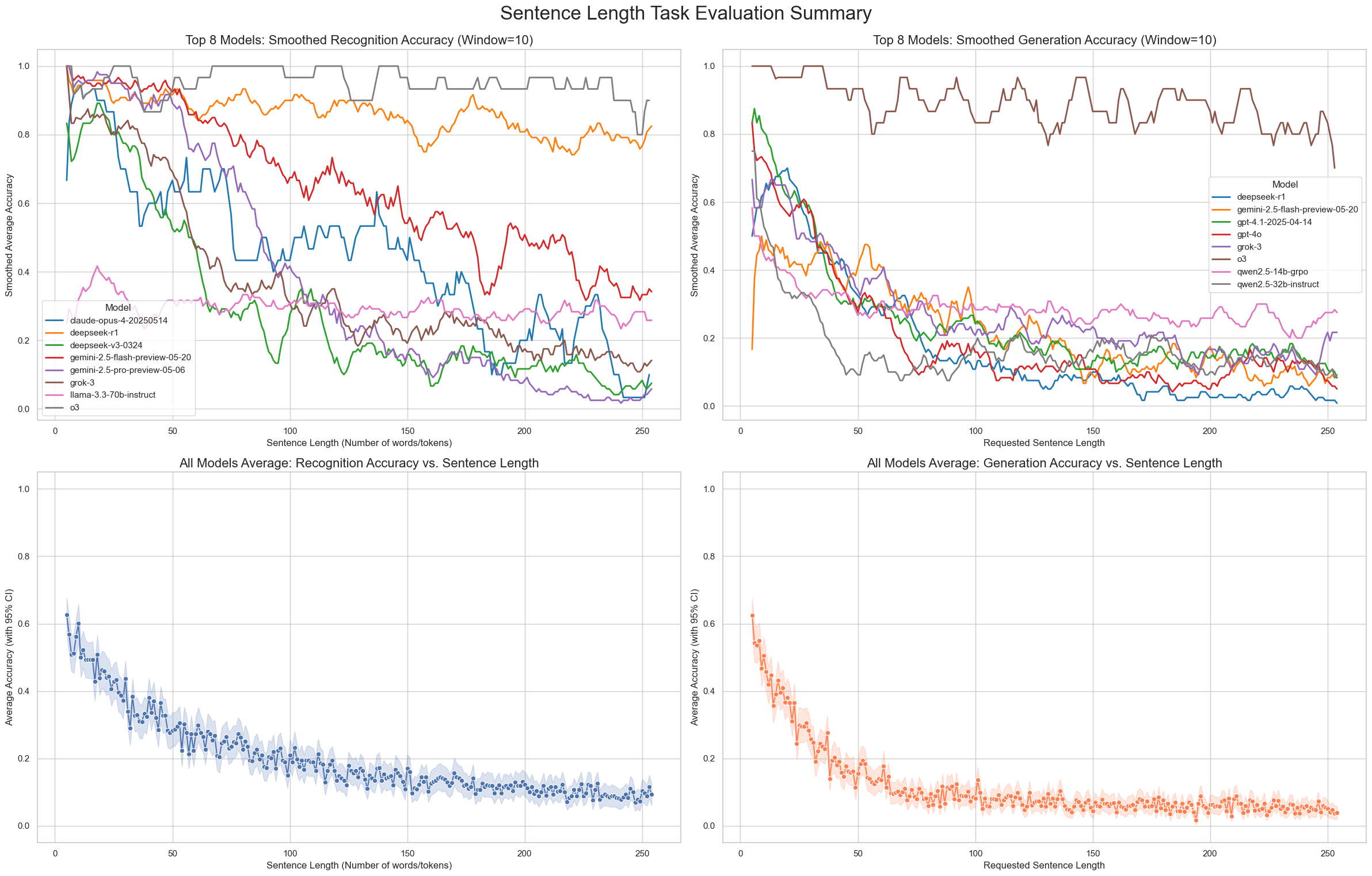}
    \caption{Summary of Sentence Length Task Evaluation. \textbf{Top row:} Smoothed accuracy for top 8 models on recognition (left) and generation (right) tasks. \textbf{Bottom row:} Average accuracy for all models on recognition (left) and generation (right) tasks.}
    \label{fig:combined_plots}
\end{figure} 

\subsubsection{General Performance Declines Sharply with Length}
Across all evaluated models, there is a clear and strong negative correlation between accuracy and sentence length for both recognition and generation tasks. As shown by the aggregated data in the bottom row of Figure~\ref{fig:combined_plots}, the average accuracy for both tasks plummets as the length requirement increases from a small number of tokens to a moderate one. For the generation task, the average accuracy starts at over 60\% for requests under 10 words but falls dramatically to less than 10\% for requests longer than 75 words. A nearly identical trend occurs in the recognition task, where accuracy begins above 60\% for sentences shorter than 10 words but quickly degrades to below 20\% for sentences longer than 75 words. This demonstrates that for the vast majority of LLMs, the ability to maintain precise control over token count is extremely fragile and does not scale, failing significantly even at moderate lengths.

\subsubsection{Top-Tier Models Show Significant Performance Divergence}
While the general trend is negative, a breakdown of the top-performing models reveals a stark divergence in capabilities, highlighting that this limitation is not universal, as shown in the top row of Figure~\ref{fig:combined_plots}.

\paragraph{An Exceptional Outlier} The \textbf{o3} model stands in a class of its own. In both the generation and recognition tasks, it maintains near-perfect, stable accuracy (between 80\% and 100\%) across the entire spectrum of tested lengths, from 1 to 250 words/tokens. This indicates that its architecture or training has successfully overcome the length-scaling challenge that plagues other models.

\paragraph{Degradation Among Other Leaders} Other top-tier models, while proficient at shorter lengths, still succumb to the scaling challenge. In the generation task, models like \texttt{gemini-2.5-flash-preview-05-20} and the fine-tuned \texttt{qwen2.5-14b-grpo} perform well on requests under 25 words but their accuracy decays rapidly thereafter. In the recognition task, \texttt{claude-opus-4-20250514} and \texttt{deepseek-r1} also demonstrate strong initial performance but see their accuracy steadily decline as sentence length increases, with \texttt{claude-opus-4} showing slightly more resilience at medium lengths before falling. This divergence shows that while most models' token awareness is brittle, it is not an unsolvable problem.

\subsection{In-depth Analysis of the Dot-Matrix Recognition (DOT) Task}

To further investigate the ``tokenizer blindness'' hypothesis, we conducted a detailed analysis of model performance on the Dot-Matrix Recognition (DOT) task. This task requires models to identify a character solely from its visual representation as a binary matrix, directly probing their ability to reason about character structure without a standard tokenizer. We programmatically analyzed the outputs from all evaluated models across three distinct sub-tasks: 1) classifying a bitmap into its correct character, 2) classifying a bitmap into its character \textit{category} (e.g., ``digit'', ``hanzi''), and 3) classifying a character name into its category.

Our analysis first reveals a significant performance gap between character categories, as shown in Table~\ref{tab:dot_category_performance}. While models demonstrate a high accuracy (over 96\%) in classifying a character's name into its category (\texttt{Average Character Classification Accuracy}), their ability to recognize the character from its visual bitmap (\texttt{Average Bitmap-to-Character Recognition Accuracy}) is dramatically lower across all categories. Performance is highest for `symbol` and `digit` categories, which often have simpler and more distinct visual forms. Conversely, performance on `hangul` (Korean), `hanzi` (Chinese), and `kana` (Japanese) is exceptionally poor, with recognition accuracies of 0.1\%, 1.83\%, and 1.84\% respectively. This starkly illustrates that models are fundamentally ill-equipped to process the complex compositional structures of these writing systems when presented visually.

\begin{table}[h]
\centering
\begin{tabular}{lrrr}
\toprule
\textbf{Cat.} & \textbf{Char. Acc.} & \textbf{B2C Acc.} & \textbf{B2Ch Acc.} \\
\midrule
digit & 1.0000 & 0.4180 & 0.2773 \\
greek & 0.9772 & 0.0146 & 0.0829 \\
hangul & 0.8699 & 0.0386 & 0.0010 \\
hanzi & 0.9924 & 0.0939 & 0.0183 \\
kana & 0.9676 & 0.0258 & 0.0184 \\
latin & 0.9737 & 0.2633 & 0.2168 \\
symbol & 0.9837 & 0.6813 & 0.1130 \\
\bottomrule
\end{tabular}
\caption{Average recognition accuracy on the DOT task by category. \textbf{Cat.} = character category, \textbf{Char. Acc.} = average character classification accuracy, \textbf{B2C Acc.} = bitmap-to-category accuracy, \textbf{B2Ch Acc.} = bitmap-to-character accuracy. Results highlight the extreme difficulty models face when recognizing bitmap characters, especially for complex scripts like Hangul, Hanzi, and Kana.}
\label{tab:dot_category_performance}
\end{table}

We then identified the most problematic characters where models could often identify the correct \textit{category} but failed to name the specific character. As detailed in Table~\ref{tab:dot_problematic_chars}, this list is dominated by punctuation and symbols. For instance, characters like \dots, \begin{CJK}{UTF8}{gbsn}，\end{CJK} (Chinese comma), \begin{CJK}{UTF8}{gbsn}：\end{CJK} (full-width colon), and \begin{CJK}{UTF8}{gbsn}、\end{CJK} were frequently misidentified, even when the model correctly recognized the bitmap as a ``symbol''. This indicates a profound lack of fine-grained visual discrimination, as models cannot differentiate between visually similar but distinct symbols.

\begin{table}[h]
\centering
\begin{tabular}{lrrr}
\toprule
\textbf{Char.} & \textbf{Fail Cnt.} & \textbf{Cat. Acc.} & \textbf{Char. Acc.} \\
\midrule
\dots & 25 & 0.9643 & 0.0714 \\
\begin{CJK}{UTF8}{gbsn}，\end{CJK} & 24 & 0.9600 & 0.0000 \\
\begin{CJK}{UTF8}{gbsn}：\end{CJK} & 24 & 0.9600 & 0.0000 \\
\begin{CJK}{UTF8}{gbsn}、\end{CJK} & 24 & 0.9600 & 0.0000 \\
--- & 24 & 0.9600 & 0.0000 \\
` & 24 & 0.8571 & 0.0000 \\
\begin{CJK}{UTF8}{gbsn}·\end{CJK} & 23 & 0.9600 & 0.0400 \\
\begin{CJK}{UTF8}{gbsn}“\end{CJK} & 23 & 0.9200 & 0.0000 \\
\_ & 21 & 0.9600 & 0.1600 \\
\begin{CJK}{UTF8}{gbsn}；\end{CJK} & 21 & 0.8400 & 0.0000 \\
" & 20 & 0.9286 & 0.2143 \\
\textasciitilde & 20 & 0.7500 & 0.0714 \\
$<$ & 19 & 0.7600 & 0.0000 \\
) & 19 & 0.8400 & 0.0800 \\
( & 19 & 0.7500 & 0.0714 \\
$>$ & 19 & 0.7600 & 0.0000 \\
\begin{CJK}{UTF8}{gbsn}（\end{CJK} & 18 & 0.7200 & 0.0000 \\
\begin{CJK}{UTF8}{gbsn}”\end{CJK} & 18 & 0.8800 & 0.1600 \\
\begin{CJK}{UTF8}{gbsn}）\end{CJK} & 18 & 0.7200 & 0.0000 \\
' & 18 & 0.9600 & 0.2400 \\
\bottomrule
\end{tabular}
\caption{Top 20 problematic characters in the DOT task. \textbf{Char.} = character, \textbf{Fail Cnt.} = recognition failure count, \textbf{Cat. Acc.} = category accuracy, \textbf{Char. Acc.} = character accuracy. These characters are often classified correctly by category but confused at the character level due to visual similarity.}
\label{tab:dot_problematic_chars}
\end{table}

Finally, an analysis of the most and least difficult characters to recognize confirms these findings. The top 20 easiest characters, listed in Table~\ref{tab:dot_easiest_chars}, are almost exclusively composed of simple, linear strokes, such as `1`, \begin{CJK}{UTF8}{gbsn}一\end{CJK}, `I`, $|$, and `H`. Their low structural complexity makes them more resilient to the information loss from bitmap conversion. Conversely, while not listed, the hardest characters to identify correspond to the Hangul and Hanzi categories, which feature intricate, multi-stroke, and spatially complex designs that are poorly captured and understood by the models in a visual modality.

\begin{table}[h]
\centering
\begin{tabular}{lr|lr}
\toprule
\textbf{Char.} & \textbf{B2Ch Acc.} & \textbf{Char.} & \textbf{B2Ch Acc.} \\
\midrule
1 & 0.8800 & A & 0.4800 \\
\begin{CJK}{UTF8}{gbsn}一\end{CJK} & 0.8400 & \begin{CJK}{UTF8}{gbsn}十\end{CJK} & 0.4800 \\
I & 0.8000 & x & 0.4800 \\
$|$ & 0.8000 & T & 0.4643 \\
H & 0.6800 & \begin{CJK}{UTF8}{gbsn}中\end{CJK} & 0.4286 \\
- & 0.5600 & \begin{CJK}{UTF8}{gbsn}二\end{CJK} & 0.4286 \\
i & 0.5600 & 8 & 0.4286 \\
$+$ & 0.5357 & \begin{CJK}{UTF8}{gbsn}口\end{CJK} & 0.4000 \\
$=$ & 0.5200 & \begin{CJK}{UTF8}{gbsn}三\end{CJK} & 0.4000 \\
: & 0.5200 & L & 0.3600 \\
\bottomrule
\end{tabular}
\caption{Top 20 easiest characters to recognize from bitmap input. \textbf{Char.} = character, \textbf{B2Ch Acc.} = bitmap-to-character accuracy. These characters tend to have simple, distinct shapes that are easily captured in visual form.}
\label{tab:dot_easiest_chars}
\end{table}

\section{Brief Introduction to GRPO}
The GRPO algorithm\cite{shao2024deepseekmath} fine-tunes large language models by replacing the value function model in conventional PPO with a group-based relative reward estimation. Instead of training a critic network to compute the advantage function, GRPO generates multiple candidate outputs for the same input, assigns each a reward score, and uses the average reward within the group as a baseline. This design not only removes the need for a separate value function but also significantly reduces memory and computation costs. Additionally, GRPO moves the KL divergence term—traditionally embedded within the GAE advantage calculation in PPO—to an explicit external regularization term with a modified computation method. Such properties make GRPO particularly well-suited for tasks with complex, rule-based evaluation metrics.

\subsection{Reward Function Design}
Our fine-tuning strategy centers on reward function design. We explore two schemes: a coarse-grained binary reward and a fine-grained partial-credit reward. This helps us assess how feedback granularity affects specialization on fine-grained tasks.

\subsubsection{Coarse-grained Reward}
The coarse-grained reward scheme adopts a binary scoring mechanism, wherein each model output is evaluated solely based on whether it is entirely correct or not. If the prediction exactly matches the ground truth according to the task’s evaluation function, it receives a full reward score of 1.0; otherwise, it receives a score of 0.0. This approach treats the task as a strict classification problem, focusing on whether the final output is completely correct without considering partial correctness or degrees of similarity.

This strategy aligns directly with many benchmark evaluation metrics that emphasize absolute correctness, such as exact match or heuristic-based binary judgments. It is particularly effective in tasks where even small deviations can significantly alter the outcome, or where partial credit may be misleading. Moreover, by reducing the reward signal to a simple success/failure indicator, this method simplifies the learning signal and allows the model to focus purely on optimizing for final accuracy.

However, while this method is straightforward and computationally efficient, it also introduces some limitations. Notably, it provides no feedback to the model when its predictions are nearly correct but slightly flawed. As a result, during training, the model might struggle to improve if its outputs are close to correct but always receive a score of 0.0, offering no guidance on how to improve further.

\subsubsection{Fine-grained Reward}
To overcome the limitations of binary feedback, we design a fine-grained reward function that provides partial credit based on the closeness of the predicted output to the correct answer. Instead of making an all-or-nothing judgment, this reward scheme evaluates how well the model performs on subtasks or dimensions of correctness, such as token order, component structure, numeric accuracy, or semantic overlap.

The fine-grained reward is particularly useful in complex tasks where outputs can be partially correct or exhibit a gradient of quality. For example, in component manipulation tasks, the reward may be based on the overlap between predicted and correct components, while in length prediction or token reordering tasks, the score reflects how far the prediction deviates from the target quantitatively. This provides the model with a more informative and differentiable reward landscape, allowing for incremental improvements during fine-tuning.

From a training perspective, fine-grained rewards offer smoother optimization signals, especially when the output space is large or discrete. They encourage the model to learn by degrees, rewarding it for producing outputs that are “on the right track” even if not perfect. This often leads to better convergence and robustness, particularly in tasks that require structured or compositional reasoning.

The implementation of our fine-grained reward scheme is encapsulated in Algorithm~\ref{alg:fine_grained_reward}, which covers multiple evaluation types including binary matching, structural similarity, length deviation, token shuffling, and numeric precision. Each case is handled with tailored heuristics or scoring formulas to ensure that the reward reflects the specific challenges of the task at hand.

\begin{algorithm}
\caption{Fine-grained Reward Calculation}
\label{alg:fine_grained_reward}
\begin{algorithmic}[1]
\Procedure{GetReward}{$t, p, l, q$}
    \State $R \gets 0$
    \If{$t$ is MATCH\_ANSWER or DIFF}
        \State $R \gets \Call{EvalBin}{p, l, q}$
    \ElsIf{$t$ is SPLIT}
        \State $P_p \gets \Call{GetParts}{p},\; O \gets \Call{GetOpts}{l}$
        \For{$o$ in $O$}
            \State $P_o \gets \Call{GetParts}{o}$
            \State $R \gets \max\left(R, \frac{|P_p \cap P_o|}{|P_o|}\right)$
        \EndFor
    \ElsIf{$t$ is SHUFFLE}
        \State $T_1, T_2 \gets \Call{Tok}{q},\; \Call{Tok}{p}$
        \If{$\Call{Match}{T_1, T_2}$}
            \State $N \gets |T_1|-1,\; A \gets \Call{AdjCnt}{T_1, T_2}$
            \State $R \gets (N > 0)\ ?\ 1 - A/N : 1$
        \EndIf
    \ElsIf{$t$ is LENGTH}
        \State $L_t \gets \Call{LenT}{q},\; L_p \gets \Call{LenP}{p}$
        \State $R \gets (L_t > 0)\ ?\ \max(0, 1 - |L_p - L_t| / L_t) : (L_p == 0\ ?\ 1 : 0)$
    \ElsIf{$t$ is NUMBER}
        \State $N_l \gets \Call{LblNum}{l},\; N_p \gets \Call{PredNum}{p}$
        \State $R \gets 1 / (1 + (N_l - N_p)^2)$
    \EndIf
    \State \Return $R$
\EndProcedure
\end{algorithmic}
\end{algorithm}

\subsection{Reward Formulas}
We summarize below the mathematical formulations used for different evaluation types. Each formula is tailored to the structure and evaluation goal of the specific task.

\begin{itemize}
  \item \textbf{Length (LENOP)}:  
  This reward penalizes deviations from the target length. It linearly decays with the relative absolute error between predicted length $L_p$ and target length $L_t$.
  \[
  R = \max\left(0,\ 1 - \frac{|L_p - L_t|}{L_t} \right)
  \]

  \item \textbf{Reorder (REORD)}:  
  This metric evaluates token reordering. A strict prerequisite is that the set of tokens must match; otherwise, the reward is zero. If satisfied, the score decreases based on how many original adjacent pairs remain.
  \[
  R =
  \begin{cases}
    1 - \frac{N_a}{|T|-1}, & \text{if token sets match} \\
    0, & \text{otherwise}
  \end{cases}
  \]

  \item \textbf{Component Match (COMPM)}:  
  For structure decomposition tasks with multiple valid ground truths, the reward is computed as the maximum overlap ratio between predicted and any reference component sets.
  \[
  R = \max_{o \in \text{Opts}} \left( \frac{|\text{Parts}_p \cap \text{Parts}_o|}{|\text{Parts}_o|} \right)
  \]

  \item \textbf{Count (FREQ/COMPC)}:  
  This formula assigns partial credit for numerical prediction tasks. It follows a squared error decay, penalizing larger deviations more harshly.
  \[
  R = \frac{1}{1 + (N_l - N_p)^2}
  \]

  \item \textbf{Exact Match (DIFF/others)}:  
  For tasks that require strict match or heuristic judgment, a binary reward is used: 1 for exact or acceptable match, 0 otherwise.
  \[
  R = \mathbb{I}(\text{valid match})
  \]
\end{itemize}

\subsection{Training Details}
We fine-tuned the \texttt{qwen2.5-14b-instruct} model using synthetic data from TASE. Table~\ref{tab:hyperparams} shows the selected hyperparameters based on preliminary stability tests.

\begin{table}[h!]
\centering
\begin{tabular}{lc}
\toprule
\textbf{Hyperparameter} & \textbf{Value} \\
\midrule
Finetuning Type & Full \\
Base Model & Qwen2.5-14B-Instruct \\
Flash Attention & FlashAttention v2 \\
GRPO Async with vLLM & True \\
Max Input Length & 16384 \\
\midrule
Batch Size (per device) & 4 \\
Gradient Accum. Steps & 4 \\
Effective Batch Size & 16 \\
Learning Rate & $3 \times 10^{-6}$ \\
Epochs & 5 \\
Scheduler & Cosine \\
Warmup Ratio & 0.1 \\
Seed & 42 \\
\midrule
Precision & bfloat16 (bf16) \\
DeepSpeed Config & ZeRO Stage 3 \\
Use Liger Kernel & True \\
Cache Overwrite & True \\
Preproc. Workers & 16 \\
Save Steps & 25 \\
Logging Steps & 1 \\
\bottomrule
\end{tabular}
\caption{Hyperparameters for GRPO Fine-tuning}
\label{tab:hyperparams}
\end{table}
\subsection{Training Data}
We constructed the GRPO training set using a synthetic data generation pipeline tailored for the TASE benchmark. For each question instance, we generated 500 new examples to ensure sufficient diversity and coverage across the task distribution. These examples were created programmatically based on task-specific rules or transformations, preserving the original problem structure while introducing variation in input-output patterns.

The resulting training set contains a mixture of instances across all TASE task types. Each training example is annotated with a \texttt{type} field in the JSON data format, indicating the corresponding task category (e.g., MATCH\_ANSWER, SPLIT, SHUFFLE, LENGTH, NUMBER). During GRPO fine-tuning, this type annotation is used to dynamically select the appropriate reward function as described in the fine-grained reward algorithm (Algorithm~\ref{alg:fine_grained_reward}).

This multi-type training setup allows for mixed-task training, where the model simultaneously learns from multiple reward functions within a single training run. Following training, we perform full evaluation on the held-out validation set to assess generalization across all task types.



\section{Case Study: From Diagnosis to Direction}

To provide a more concrete illustration of the models' strengths and weaknesses, this section presents a case study of specific examples from the TASE benchmark. These cases are selected not only to highlight the core challenges discussed in this paper---particularly the gap in visual and structural understanding---but also to connect these diagnostic findings to potential pathways for improvement. By examining model failures in detail, and linking them to the quantitative results in this paper, we can better appreciate the value of the mitigation strategies explored and the broader research directions they represent.

\subsection{Case 1: Dot-Matrix Recognition and Visual Reasoning Limits}

The Dot-Matrix Recognition (DOT) task evaluates a model's ability to interpret a character's visual form from a low-level pixel representation, a skill that requires spatial reasoning beyond simple text processing. This task probes whether a model can genuinely "see" and classify a shape.

\textbf{Task:} Classify a 16x16 bitmap into the symbol it represents.

\textbf{Input:}
\begin{tcolorbox}
\begin{quote}
Please classify the following 16x16 bitmap into one of the following
categories: [...] symbol (punctuation or other symbols)
bitmap:
(matrix for '!')
\end{quote}
\end{tcolorbox}

\textbf{Correct Answer:}
\begin{tcolorbox}
\begin{quote}
symbol
\end{quote}
\end{tcolorbox}

\subsubsection{Model Performance}

\textbf{Correct Prediction (claude-sonnet-4-20250514):}
\begin{tcolorbox}
\begin{quote}
Looking at the bitmap pattern, I can see a straight vertical line in the
top half and a small dot/period at the bottom. This pattern clearly
represents an exclamation mark (!), which belongs in the symbol category.
\texttt{<answer>}symbol\texttt{</answer>}
\end{quote}
\end{tcolorbox}

\textbf{Incorrect Prediction (claude-3-7-sonnet-20250219):}
\begin{tcolorbox}
\begin{quote}
Looking at the bitmap pattern, I can see it shows a vertical line with a
small filled rectangle at the bottom - this appears to be the lowercase
letter 'i' which consists of a vertical stroke and a dot. Therefore,
this bitmap represents a Latin letter.
\texttt{<answer>}latin\texttt{</answer>}
\end{quote}
\end{tcolorbox}

\subsubsection{Analysis}
This task requires two levels of abstraction: first, recognizing the shape from the pixel grid, and second, mapping that shape to a linguistic category. The successful model, \texttt{claude-sonnet-4}, which scores 41.67\% on the DOT task (Table \ref{tab:tase-results}), correctly identifies the shape and classifies it. In contrast, the incorrect model, \texttt{claude-3-7-sonnet-20250219} (39.58\% on DOT), makes a subtle but critical error. It correctly perceives the visual components but incorrectly associates the pattern with the letter 'i'. This reveals a critical weakness: the model's reasoning is not robust. It has learned a visual association (line + dot) but cannot reliably distinguish its usage in '!' versus 'i'. This supports the paper's conclusion that models often lack deep structural discrimination, relying instead on shallow pattern matching.

\subsection{Case 2: Component Splitting and Chinese Structural Deficits}
This case, from the Component Manipulation (COMPM) family, assesses the model's understanding of the compositional nature of logographic characters, a key aspect of structural understanding in languages like Chinese.

\textbf{Task:} 
\begin{CJK}{UTF8}{gbsn}
将汉字“究”拆分为可重新组合的基本部件。
\end{CJK}

\textbf{Input:}
\begin{tcolorbox}
\begin{CJK}{UTF8}{gbsn}
\begin{quote}
请将“究”拆分为可重新组合的基本部件.
\end{quote}
\end{CJK}
\end{tcolorbox}

\textbf{Correct Answer:}
\begin{tcolorbox}
\begin{CJK}{UTF8}{gbsn}
\begin{quote}
穴,九
\end{quote}
\end{CJK}
\end{tcolorbox}

\subsubsection{Model Performance}

\textbf{Correct Prediction (claude-sonnet-4-20250514):}
\begin{tcolorbox}
\begin{CJK}{UTF8}{gbsn}
\begin{quote}
"究" 可以拆分为以下基本部件：
\vspace{1mm}

穴 + 九
\vspace{1mm}

\texttt{<answer>}穴 + 九\texttt{</answer>}
\end{quote}
\end{CJK}
\end{tcolorbox}

\textbf{Incorrect Prediction (claude-3-7-sonnet-20250219):}
\begin{tcolorbox}
\begin{CJK}{UTF8}{gbsn}
\begin{quote}
“究”字的基本部件可以拆分为：
\vspace{1mm}

1. **小**：表示“小”的意思。
2. **糸**（米字旁）：表示与“丝”或“线”相关的事物。
\end{quote}
\end{CJK}
\end{tcolorbox}

\subsubsection{Analysis}
This example highlights the challenges LLMs face with the structural properties of non-alphabetic languages. The character \begin{CJK}{UTF8}{gbsn}究\end{CJK} (jiū) is composed of the radical \begin{CJK}{UTF8}{gbsn}穴\end{CJK} (xuè, cave) and the component \begin{CJK}{UTF8}{gbsn}九\end{CJK} (jiǔ, nine). The successful model, \texttt{claude-sonnet-4} (45.41\% on Chinese tasks), correctly identifies these components. The incorrect model, \texttt{claude-3-7-sonnet-20250219} (41.59\% on Chinese tasks), provides a completely erroneous decomposition. This failure is a stark demonstration of a lack of structural understanding. The model does not possess a coherent model of how Chinese characters are constructed, instead hallucinating a plausible-sounding but incorrect answer, likely due to faulty associations learned during training.

\subsection{Case 3: Component Combination and Cross-Lingual Structural Deficits}
This case also comes from the COMPM family, testing the model's ability to apply compositional rules for Korean Hangul, where syllables are systematically constructed from Jamo components.

\textbf{Task:} 
\begin{CJK}{UTF8}{mj}
다음 자모를 조합하세요 (Combine the following Jamo).
\end{CJK}

\textbf{Input:}
\begin{tcolorbox}
\begin{CJK}{UTF8}{mj}
\begin{quote}
다음 자모를 조합하세요: ㅆ, ㅗ, ㄹ
\end{quote}
\end{CJK}
\end{tcolorbox}

\textbf{Correct Answer:}
\begin{tcolorbox}
\begin{CJK}{UTF8}{mj}
\begin{quote}
쏠
\end{quote}
\end{CJK}
\end{tcolorbox}

\subsubsection{Model Performance}

\textbf{Correct Prediction (claude-opus-4-20250514):}
\begin{tcolorbox}
\begin{CJK}{UTF8}{mj}
\begin{quote}
주어진 자모 ㅆ, ㅗ, ㄹ을 조합하여 한글을 만들어보겠습니다.
ㅆ (초성) + ㅗ (중성) + ㄹ (종성) = 쏠
\texttt{<answer>}쏠\texttt{</answer>}
\end{quote}
\end{CJK}
\end{tcolorbox}

\textbf{Incorrect Prediction (doubao-1-5-lite-32k-250115):}
\begin{tcolorbox}
\begin{CJK}{UTF8}{mj}
\begin{quote}
\texttt{<answer>}실례합니다, ㅆ, ㅗ, ㄹ을 조합할 수 있는 자음과 모음의 조합은 없습니다.
ㅆ는 자음이며, ㅗ와 ㄹ도 자음입니다. 자음과 자음을 조합할 수는 없습니다.
[...]\texttt{</answer>}
\end{quote}
\end{CJK}
\end{tcolorbox}

\subsubsection{Analysis}
This example starkly demonstrates the structural understanding gap in non-English languages. The successful model, \texttt{claude-opus-4}, correctly identifies the roles of each Jamo and combines them. Yet, its overall performance on Korean tasks is only 36.19\%, showing that even success on one example does not imply mastery. The incorrect model, \texttt{doubao-lite}, whose Korean performance is a very low 10.65\%, fails catastrophically. It incorrectly identifies the vowel \begin{CJK}{UTF8}{mj}ㅗ\end{CJK} as a consonant and then falsely concludes that the combination is impossible. This is not just a simple error but a complete failure to understand the basic structural principles of the language, a failure directly reflected in its low quantitative score.

\subsection{Direct Impact of GRPO Fine-Tuning: A Comparative Case Study}
The most direct way to validate the effectiveness of the TASE benchmark and the GRPO training method is to observe the performance changes of the same base model before and after fine-tuning. The following cases compare the `qwen2.5-14b-instruct` model with its GRPO-tuned version, `qwen2.5-14b-grpo`, on tasks where the base model initially failed.

\subsubsection{Case 4: GRPO's Effect on Structural Understanding (Variant Normalization)}
This case demonstrates how GRPO fine-tuning enhances the model's ability to understand and restore structurally complex and visually similar Chinese characters.

\textbf{Task:} 
\begin{CJK*}{UTF8}{gbsn}
还原被视觉混淆的中文句子。
\end{CJK*}

\textbf{Input:}
\begin{tcolorbox}
\begin{CJK*}{UTF8}{gbsn}
\begin{quote}
以下是被扰动后的文本，请你还原出原始文本，不修改标点符号：导湮湜電影狆哋募後渶雄
\end{quote}
\end{CJK*}
\end{tcolorbox}

\textbf{Correct Answer:}
\begin{tcolorbox}
\begin{CJK*}{UTF8}{gbsn}
\begin{quote}
导演是电影中的幕后英雄
\end{quote}
\end{CJK*}
\end{tcolorbox}

\paragraph{Model Performance Comparison}

\textbf{Prediction (qwen2.5-14b-instruct, Pre-GRPO):}
\begin{tcolorbox}
\begin{CJK*}{UTF8}{gbsn}
\begin{quote}
\texttt{<answer>}导入湜电影中的幕后功\texttt{</answer>}
\end{quote}
\end{CJK*}
\end{tcolorbox}

\textbf{Prediction (qwen2.5-14b-grpo, Post-GRPO):}
\begin{tcolorbox}
\begin{CJK*}{UTF8}{gbsn}
\begin{quote}
\texttt{<answer>}导演是电影中的幕后英雄\texttt{</answer>}
\end{quote}
\end{CJK*}
\end{tcolorbox}

\paragraph{Analysis}
The base instruct model demonstrates limited robustness, often reverting to frequent but incorrect token sequences, suggesting a superficial grasp of structural patterns. In contrast, the grpo model consistently produces accurate outputs, indicating that fine-tuning with TASE's structurally diverse synthetic data effectively equips the model with the contextual and visual sensitivity required to mitigate tokenizer-related deficiencies.

\subsubsection{Case 5: GRPO's Effect on Token Awareness (Token Shuffling)}
This case shows how GRPO improves the model's fundamental ability to "see" and manipulate individual tokens as discrete units, a core aspect of token awareness.

\textbf{Task:} 
\begin{CJK*}{UTF8}{gbsn}
完全打乱句子中的字。
\end{CJK*}

\textbf{Input:}
\begin{tcolorbox}
\begin{CJK*}{UTF8}{gbsn}
\begin{quote}
请完全打乱下面这些字，确保每个字的前后位置都与原句不相邻。
输入：除此之外的其他城市其
\end{quote}
\end{CJK*}
\end{tcolorbox}

\paragraph{Model Performance Comparison}

\textbf{Prediction (qwen2.5-14b-instruct, Pre-GRPO):}
\begin{tcolorbox}
\begin{CJK*}{UTF8}{gbsn}
\begin{quote}
\texttt{<answer>}城市之外的此其其他\texttt{</answer>}
\end{quote}
\end{CJK*}
\end{tcolorbox}

\textbf{Prediction (qwen2.5-14b-grpo, Post-GRPO):}
\begin{tcolorbox}
\begin{CJK*}{UTF8}{gbsn}
\begin{quote}
首先，我们将输入的字符串分解成单独的汉字："此" "之" "外" "的" "其" "他" "城" "市" "其"
然后，我们随机打乱这些汉字的位置，得到一个新序列："城" "其" "市" "之" "的" "其" "外" "他" "此"
所以，答案为：\texttt{<answer>}城其市之的其外他此\texttt{</answer>}
\end{quote}
\end{CJK*}
\end{tcolorbox}

\paragraph{Analysis}
The base model provides a poor-quality shuffle, with several words remaining in close proximity to their original positions. The `grpo` model, however, not only produces a much more thoroughly randomized sequence but also exhibits Chain-of-Thought reasoning by explaining its process of splitting and shuffling the characters. This demonstrates a much deeper token awareness. The GRPO training has not only improved the model's ability to perform the task but has also enhanced its ability to reason about the procedural steps involved, a significant upgrade in capability.

\subsection{The Power of Reasoning: A Chain-of-Thought Case Study}
Beyond fine-tuning, another key method for improving model performance is enhancing the reasoning process at inference time. Chain-of-Thought (CoT) prompting encourages the model to "think step-by-step," breaking down a problem before reaching a conclusion. This case study demonstrates its impact on a `difftokens` task, which requires careful comparison and attention to detail.

\textbf{Task:} Are the words in seq1 and seq2 exactly matching one-to-one (ignoring order)? If yes, answer 'yes'. If not, which word is different?

\textbf{Input:}
\begin{tcolorbox}
\begin{quote}
seq1: he good Although almost started \\
seq2: only almost he good started Although
\end{quote}
\end{tcolorbox}

\textbf{Correct Answer:}
\begin{tcolorbox}
\begin{quote}
only
\end{quote}
\end{tcolorbox}

\subsubsection{Model Performance Comparison}

\textbf{Prediction (gpt-3.5-turbo, Standard Prompting):}
\begin{tcolorbox}
\begin{quote}
\texttt{<answer>}no, the word "only" is different\texttt{</answer>}
\end{quote}
\end{tcolorbox}

\textbf{Prediction (gpt-3.5-turbo, CoT Prompting):}
\begin{tcolorbox}
\begin{quote}
To determine if the words in seq1 and seq2 match one-to-one while ignoring order, we can follow these steps:

1.  **Identify the words in both sequences:**
    \begin{itemize}
        \item seq1: he, good, Although, almost, started
        \item seq2: only, almost, he, good, started, Although
    \end{itemize}
2.  **Count the occurrences of each word in both sequences...**
3.  **Compare the counts:**
    \begin{itemize}
        \item The words "he", "good", "Although", "almost", and "started" appear in both sequences with the same frequency.
        \item However, seq2 contains the word "only", which does not appear in seq1.
    \end{itemize}
The different word is "only".

\texttt{<answer>}only\texttt{</answer>}
\end{quote}
\end{tcolorbox}

\subsubsection{Analysis}
This is a classic example of how CoT enhances not just accuracy but also instruction following and reliability. The base `gpt-3.5-turbo` model correctly identifies the different word ("only"), but it fails to follow the output format instruction, embedding the answer in a natural language sentence instead of the required ` <answer> ` tag. This is a subtle but common failure mode.

The `gpt-3.5-turbo` model with cot, in contrast, follows a perfect, explicit reasoning path. It lists the words, compares them, identifies the discrepancy, and then, as a final step, formats its conclusion precisely as requested. This structured approach prevents the model from taking shortcuts and ensures that all constraints of the prompt, including complex formatting rules, are met. This demonstrates that CoT is a powerful tool for turning a "smart" but sometimes unreliable model into a "diligent" and precise one.

\subsection{The Power of Reasoning: A Chain-of-Thought Case Study}
Beyond fine-tuning, another key method for improving model performance is enhancing the reasoning process at inference time. Chain-of-Thought (CoT) prompting encourages the model to "think step-by-step," breaking down a problem before reaching a conclusion. This case study demonstrates its impact on a `componentcount` task, which requires careful counting and attention to detail.

\textbf{Task}: How many times does the letter "p" appear in rope-a-dope polyposis all-optical?

\textbf{Input:}
\begin{tcolorbox}
\begin{quote}
How many times does the letter "p" appear in rope-a-dope polyposis all-optical?
\end{quote}
\end{tcolorbox}

\textbf{Correct Answer:}
\begin{tcolorbox}
\begin{quote}
5
\end{quote}
\end{tcolorbox}

\subsubsection{Model Performance Comparison}

\textbf{Prediction (gpt-3.5-turbo, Standard Prompting):}
\begin{tcolorbox}
\begin{quote}
\texttt{<answer>}6\texttt{</answer>} 
\end{quote}
\end{tcolorbox}

\textbf{Prediction (gpt-3.5-turbo, CoT Prompting):}
\begin{tcolorbox}
\begin{quote}
To find out how many times the letter "p" appears in the phrase "rope-a-dope polyposis all-optical," let's break it down word by word:

1. **rope-a-dope**: The letter "p" appears 2 times (in "rope" and "dope"). 
2. **polyposis**: The letter "p" appears 2 times. 
3. **all-optical**: The letter "p" appears 1 time. 

Now, let's sum these occurrences:

- From "rope-a-dope": 2
- From "polyposis": 2
- From "all-optical": 1

Total occurrences of "p": 2 + 2 + 1 = 5 

Thus, the letter "p" appears 5 times in the phrase "rope-a-dope polyposis all-optical."

\texttt{<answer>}5\texttt{</answer>}
\end{quote}
\end{tcolorbox}

\subsubsection{Analysis}
This is a powerful illustration of how CoT mitigates "careless" errors. The task is not conceptually difficult, yet the standard model fails. Its reasoning shows a clear misstep: it incorrectly counts three 'p's in "polyposis" instead of two. This is a classic attentional failure, where the model processes the input too quickly and makes a simple counting mistake.

The CoT-prompted model, however, succeeds by adopting a more rigorous, self-correcting procedure. By explicitly breaking the problem down "word by word," it forces a more granular level of attention. This structured approach allows it to correctly count the letters in each segment ("rope-a-dope": 2, "polyposis": 2, "all-optical": 1) and arrive at the correct sum. This case demonstrates that CoT is not just for complex, multi-step logical problems; it is also a crucial tool for improving the fundamental reliability of a model on simpler tasks by enforcing a diligent and methodical process, turning a "fast but sloppy" reasoner into a "deliberate and accurate" one.

\section{Connecting Failures to Future Directions}
The weaknesses exposed in these cases are not insurmountable. The TASE benchmark itself, and the analyses it enables, point toward several promising solutions. This paper explores two of these in depth: targeted fine-tuning and advanced prompting strategies. In addition, it highlights a broader research direction worthy of future investigation.

\textbf{Targeted Fine-Tuning (GRPO):} The failures in the Chinese and Korean cases are knowledge gaps. The models do not robustly understand the rules of character formation. As shown in Section 5, fine-tuning the \texttt{qwen2.5-14b} model with the GRPO method on the TASE synthetic dataset nearly doubled its average score (from 11.72\% to 20.40\%). This demonstrates that these structural rules can be explicitly taught, allowing the model to learn the underlying linguistic patterns that its pre-training failed to provide.

\textbf{Chain-of-Thought (CoT) Prompting:} The failure in the Dot-Matrix case stems from a flawed reasoning process. The model makes a superficial leap in logic (line + dot = ‘i’). Section 6.6 shows that CoT prompting systematically improves performance on TASE tasks by forcing the model to "think step-by-step." For the DOT task, a CoT prompt would encourage the model to consider alternative interpretations of the visual pattern before jumping to a conclusion, reducing the likelihood of a premature, incorrect classification.

\textbf{Broader Research Context:} These cases also underscore the importance of research into tokenizer-free or character-aware architectures. Models like CharBERT\cite{ma2020charbert}, which learn representations directly from characters, and novel tokenization schemes that respect linguistic boundaries are critical areas of future work. The TASE benchmark serves as an essential diagnostic tool to measure progress and guide the development of next-generation models that unite high-level semantic comprehension with the low-level precision and structural awareness demonstrated to be lacking in these case studies.

\bibliography{aaai2026}

\bibliography{aaai2026}

\end{document}